\pdfoutput=1

\documentclass[11pt]{article}

\usepackage[]{ACL2023}

\usepackage{times}
\usepackage{latexsym}
\usepackage{mathtools}

\usepackage{graphicx}
\usepackage[utf8]{inputenc}
\newcommand\textcyr[1]{{\fontencoding{T2A}\selectfont #1}}
\newcommand\comment[1]{}

\newcommand{\CUEW}[1]{{\color{darkblue} {#1}}}

\usepackage{microtype}
\usepackage{amsmath}
\usepackage{amsfonts}
\usepackage{amssymb}
\usepackage{algorithm}
\usepackage{algorithmic}
\usepackage{bm}

\usepackage{inconsolata}
\usepackage[russian,french,english]{babel}
\usepackage[utf8]{inputenc}
\usepackage[T2A,OT1]{fontenc}
\usepackage{tempora} 

\newcommand{\VSPACEPAR}[1]{\vspace{0.1cm}
\noindent \textbf{#1} \ }

\newenvironment{itemizerCompact}{\vspace{-1mm}
  \begin{itemize}
    \setlength{\itemsep}{1pt}
    \setlength{\parskip}{0pt}
    \setlength{\parsep}{0pt}
  }
{ \end{itemize}
  \vspace{-1mm}  }

\newenvironment{enumeratorCompact}{\vspace{-1mm}
  \begin{enumerate}
    \setlength{\itemsep}{2pt}
    \setlength{\parskip}{0pt}
    \setlength{\parsep}{0pt}
  }
{ \end{enumerate}
  \vspace{-1mm}  }

\title{Probing the Category of Verbal Aspect in Transformer Language Models}

\author{Anisia Katinskaia,\textsuperscript{*$\diamondsuit$} \ 
  Roman Yangarber\textsuperscript{$\diamondsuit$} \\
  \textsuperscript{*} Department of Computer Science \\
  \textsuperscript{$\diamondsuit$} Department of Digital Humanities \\
  University of Helsinki, Finland \\
  \texttt{first.last@helsinki.fi}}

\begin{document}
\maketitle
\begin{abstract}
We investigate how pretrained language models (PLM) encode the grammatical category of verbal aspect in Russian.  Encoding of aspect in transformer LMs has not been studied previously in any language. 
A particular challenge is posed by {\em ``alternative contexts''}: where either the perfective or the imperfective aspect is suitable grammatically and semantically.  We perform probing using BERT and RoBERTa on alternative and non-alternative contexts.  First, we assess the models' performance on aspect prediction, via behavioral probing.  Next, we examine the models' performance when their contextual representations are substituted with counterfactual representations, via causal probing.  These counterfactuals alter the value of the ``boundedness'' feature---a semantic feature, which characterizes the action in the context.  
Experiments show that BERT and RoBERTa do encode aspect---mostly in their final layers.
The counterfactual interventions affect perfective and imperfective in opposite ways, which is consistent with grammar: perfective is positively affected by adding the meaning of boundedness, and vice versa.  
\comment{??? <Q1. is this CRUCIAL enough to include in abstract?  Q2. if it is, it should be near the final statement -- also about alt. contexts> Alternative contexts appear to be more sensitive to counterfactual interventions. Answer: No}  
The practical implications of our probing results are that fine-tuning only the last layers of BERT on predicting aspect is faster and more effective than fine-tuning the whole model.  
The model has high predictive uncertainty about aspect in alternative contexts, which tend to lack explicit hints about the boundedness of the described action.    

\end{abstract}

\section{Introduction}

This paper focuses on the grammatical category of {\em verbal aspect}.  It is a category that involves both morphology and semantics of the verb, and expresses how an action denoted by the verb extends over time.  
Linguistic theory of aspect is intricate: different languages make different aspectual distinctions, e.g., languages can have distinct perfective/imperfective/progressive categories of aspect, while some make no distinction at all.  Aspect is also one of the most complex categories in many languages: even advanced experts, who are non-native speakers, continue to make errors in the choice of aspect~\cite{forsyth1970grammar,bar2008selective}.  We focus on the Slavic aspectual system, in particular in Russian, which displays significant differences in the semantics of the perfective/imperfective opposition to other languages~\cite{dahl85}.  

How aspect is encoded in pretrained language models (PLMs) has not been previously studied for any language, although other grammatical properties---number agreement, predicate-argument syntactic relations, etc.---have been studied.
It is challenging to identify what linguistic phenomena in the context affect the choice of aspect. A special challenge concerning aspect is posed by {\em ``alternative contexts''}, where more than one aspect form is acceptable grammatically and semantically.


We investigate the following research questions:
RQ1. Do BERT and RoBERTa encode the category of aspect, and if they do---how? 
RQ2. How does the encoding of aspect in these models correspond to linguistic theory of aspect?  
RQ3. Is encoding of aspect in alternative contexts different from non-alternative contexts? 

We perform two kinds of probing: behavioral and causal.  In behavioral probing, we inspect layers one by one, observing how the model predicts which aspect form best suits the context.  If the model fails, we infer that it does not encode the target linguistic property (here---aspect).  We introduce two types of behavioral probing via filling a mask: iterative masking and aspect inference. 
In both methods, the model's preference for aspect is reflected in the probabilities it assigns to verb forms in the masked position.  For causal probing, we {\em intervene} in the model's representations at each layer:
we manipulate the semantics of the action described by the target verb and its context---whether the action is {\em bounded} or {\em unbounded}.
If the intervention is relevant for predicting the target property, the model's
performance on the task will be affected.  

Our findings can be summarised as follows:
(1) All probing methods indicate that BERT and RoBERTa do encode aspect, predominantly in their final layers. 
(2) Interventions in sentence semantics cause effects consistent with theory of aspect: imperfective verbs typically describe {\em unbounded} actions, while perfective verbs describe {\em bounded} actions. 
(3) Fine-tuning only the final layers of BERT for aspect prediction results in improved performance, confirming our first finding.
(4) Both pretrained and fine-tuned models exhibit {\em high uncertainty} regarding aspect preference in {\em alternative} contexts, where multiple aspect forms are valid.
(5) {\em Alternative} contexts are more sensitive to causal intervention in the semantics of boundedness.
(6) Such contexts often lack explicit hints about the action's boundedness, which makes both humans and PLMs uncertain about the choice of aspect.


\section{Related Work}

Several studies focus on the internal representation of linguistic information inside PLMs. \textit{Correlation probing} methods are based on \textit{parametric probes}, i.e., linear or non-linear classifiers trained on model representations to predict specific linguistic properties~\cite{adi2017fine, conneau-etal-2018-cram,tenney2018what,hewitt-manning-2019-structural,10.1609/aaai.v33i01.33016309, hall-maudslay-etal-2020-tale, weissweiler-etal-2022-better, conia-navigli-2022-probing, arps-etal-2022-probing}. 
Some have questioned the efficacy of probing classifiers, and whether the original model, which was used as an encoder, actually uses\comment{??? :/ ... i've not read the papers - but this sounds like a strange claim: why would a model STORE the information if it has NO USE for it? Answer: that is what papers criticize} the information discovered by probes~\cite{hewitt-liang-2019-designing,tamkin-etal-2020-investigating,ravichander-etal-2021-probing}.
In response to this criticism, a number of 
methodologies are proposed~\cite{hewitt-etal-2021-conditional,pimentel-etal-2020-information,voita-titov-2020-information,immer-etal-2022-probing,wang2023gpp}. \citet{belinkov-2022-probing} gives an extensive review of probing classifiers as an approach, their advantages and shortcomings.

\textit{Non-parametric} correlation probing, or \textit{behavioral probing}, tests the behavior of PLMs without additional classifiers.  To isolate the target linguistic property, a PLM is evaluated by using a set of carefully designed examples~\cite{linzen-etal-2016-assessing, gulordava-etal-2018-colorless, ribeiro-etal-2020-beyond,Warstadt2020CanNN,newman-etal-2021-refining,wu-etal-2020-perturbed,10.1162/tacl_a_00531,10.1162/tacl_a_00554,kim-etal-2023-reconstruction}.
\citet{ravfogel-etal-2019-studying} propose a methodology for creating synthetic examples, which differ by various linguistic properties.
While most work focuses on English, ~\citet{mueller-etal-2020-cross} introduce the CLAMS dataset for syntactic evaluation of models for five languages, including Russian.
\citet{hlavnova-ruder-2023-empowering} propose Multilingual Morphological Checklist (M2C), a framework for behavioral probing of typological features in 12 languages, e.g., motion verbs in Russian.

\textit{Causal probing} relies on controlled interventions into the LM's internal components (or into the input), and studying consequent changes in the model's behavior~\cite{giulianelli-etal-2018-hood,DBLP:journals/corr/abs-2004-12265,10.1162/tacl_a_00359,Kaushik2020Learning,geiger2021causal,voita-etal-2021-analyzing,finlayson-etal-2021-causal,lasri-etal-2022-probing,rozanova-etal-2023-interventional,yamakoshi-etal-2023-causal,10.1162/tacl_a_00531}. 
{\em Amnesic} probing~\cite{10.1162/tacl_a_00359} builds on the intuition that removing a property from the representation will weaken the model's ability to solve a task, if the property is important for the task.  The approach is based on an algorithm---Iterative Null-space Projection (INLP)---for removing linear information from representations~\cite{ravfogel-etal-2020-null}. \citet{ravfogel-etal-2021-counterfactual} apply INLP to generate
counterfactual representations and use these to test how changing particular linguistic features affects the model’s behavior.
Despite some criticism of INLP~\cite{kumar2022probing}, we use it 
to investigate the behavior of LMs on aspect prediction.

\section{Background on Aspect}

The category of aspect in Russian characterizes the action described by a verb in terms of its progress---continuous vs.~punctual, completed vs.~uncompleted, etc.---or from the observer's perspective---retrospective vs.~synchronous. The meaning of aspect opposition has long been a subject of debate.  In this paper, we adhere to the theory that {\em boundedness}---reaching a limit---is the factor determining the aspect form~\cite{Vinogradov47,dahl85}. 
We assume that every verb has two aspect forms---{\em perfective} and {\em imperfective}---though in some rare cases the two forms may coincide, e.g., ``\textcyr{обещать}'' (\textit{to promise---perf. or imperf.}).      

Unlike most grammatical categories, aspect has no unique marker in the verb form and is tightly connected with the verb's lexical meaning.  Aspect can be expressed by the root, e.g., \textcyr{``\textbf{говор}ить''} (imp.) vs.~\textcyr{``\textbf{сказ}ать''} (perf.), \textit{to say}; 
by the suffix, e.g., \textcyr{``толк\textbf{а}ть''} (imp.) vs.~\textcyr{``толк\textbf{ну}ть''} (perf.), \textit{to push}; 
by the prefix, e.g., \textcyr{``делать''} (imp.) vs.~\textcyr{``\textbf{с}делать''} (perf.), \textit{to do/make}.\footnote{The prefix may affect the meaning, but lexical vs.~grammatical changes are often very difficult to disentangle; therefore we consider such verb pairs to be aspect pairs.}
Examples (1) and (2) show the aspect pair of verbs ``\textcyr{дуть / дунуть}'' (\textit{to blow}): 

\begin{enumeratorCompact}
\item \textcyr{На побережье всегда \underline{дул} (imp.) ветер.} \label{item:2}
\item[] \textit{Wind always \underline{blew} on the coast.} 
\item \textcyr{И вдруг резко \underline{дунул} (perf.) ветер.} \label{item:1}
\item[] \textit{And suddenly the wind \underline{blew} sharply.}
\end{enumeratorCompact}

In these contexts, only one aspect form is acceptable. We call such contexts \textbf{non-alternative}. However, in some (narrow) contexts it may not be possible to decide which aspect fits best, since both may fit, albeit with slight differences in meaning. We call such contexts \textbf{alternative}. For example, in the sentence below both perfective and imperfective are acceptable:

\vspace{4.5px}
    \noindent \textcyr{Я уже \underline{позвонил} (perf.) в клинику и вызвал врача.}\\
    \noindent \textcyr{Я уже \underline{\ \ \  \ звонил} (imp.) в клинику и вызвал врача.}\\
    \noindent \textit{I already \underline{rang} the clinic and called the doctor}.
    
\vspace{4.5px}
For any particular instance, we use the term {\em expected} for the original verb form found in the text vs.~the opposite form, which we call {\em complementary}.
We perform experiments by probing aspect of the expected vs.~the complementary form; we also investigate model behavior in the non-alternative vs.~alternative contexts.

\section{Experiments}

For probing experiments, we use the Russian BERT-base, BERT-large~\cite{devlin-etal-2019-bert}, and RoBERTa-large~\cite{liu2020roberta}.\footnote{\href{https://huggingface.co/ai-forever}{huggingface.co/ai-forever}}  We mostly focus on experiments with BERT-large, since other models showed similar performance. 

\subsection{Data}
There are no pre-existing datasets for probing verbal aspect, so we perform our analysis using the following data. 
For {\em alternative} contexts, we collected short paragraphs from the ReLCo corpus~\cite{katinskaia-etal-2022-semi};
the contexts contain exercises offered to learners of Russian, where they inserted verb forms that {\em differ} from the expected answers only by the aspect feature.
The learners used the Revita language teaching and learning system~\cite{katinskaia:2018-lrec:revita,katinskaia:2017-nodalida:revita}.  These forms were manually annotated as acceptable by several native speakers.  For non-alternative contexts, we created our own dataset by randomly selecting sentences from the Omnia corpus~\cite{shavrina2019omnia}.
In each context,\footnote{An instance is a long sentence or several shorter sentences.} we pick one verb (hereafter, the \textbf{target}). We generate the target verb's complementary aspect form using a morphological generator~\cite{korobov2015morphological}; further details in Appendix~\ref{sec:appendix1}.


We tried to ensure that the target verbs are lexically varied. The collected contexts with hidden target verbs and the generated aspect pairs were manually annotated by two native speakers.
The annotation task was to assess whether the given verb form fits the context grammatically and semantically.
We collected 750 non-alternative contexts---with 375 examples for each aspect---featuring 542 distinct target verb aspect pairs. 
We expanded the set of alternative contexts to 496 instances in total, with 238 perfective and 258 imperfective verbs.  The agreement between the annotators was 84.5\%, conflicts were resolved through discussion.

We release the annotated data and the first Russian Aspect Bank with over 2K unique aspect pairs with this paper.\footnote{\href{https://github.com/RevitaAI/AspectProbing}{github.com/RevitaAI/AspectProbing}} The Aspect Bank was manually created in collaboration with experts in Russian linguistics and language pedagogy.

\begin{figure*}
\begin{minipage}{0.5\linewidth}
\centering
  \includegraphics[scale=0.48, trim=0mm 0mm 0mm 0mm ]{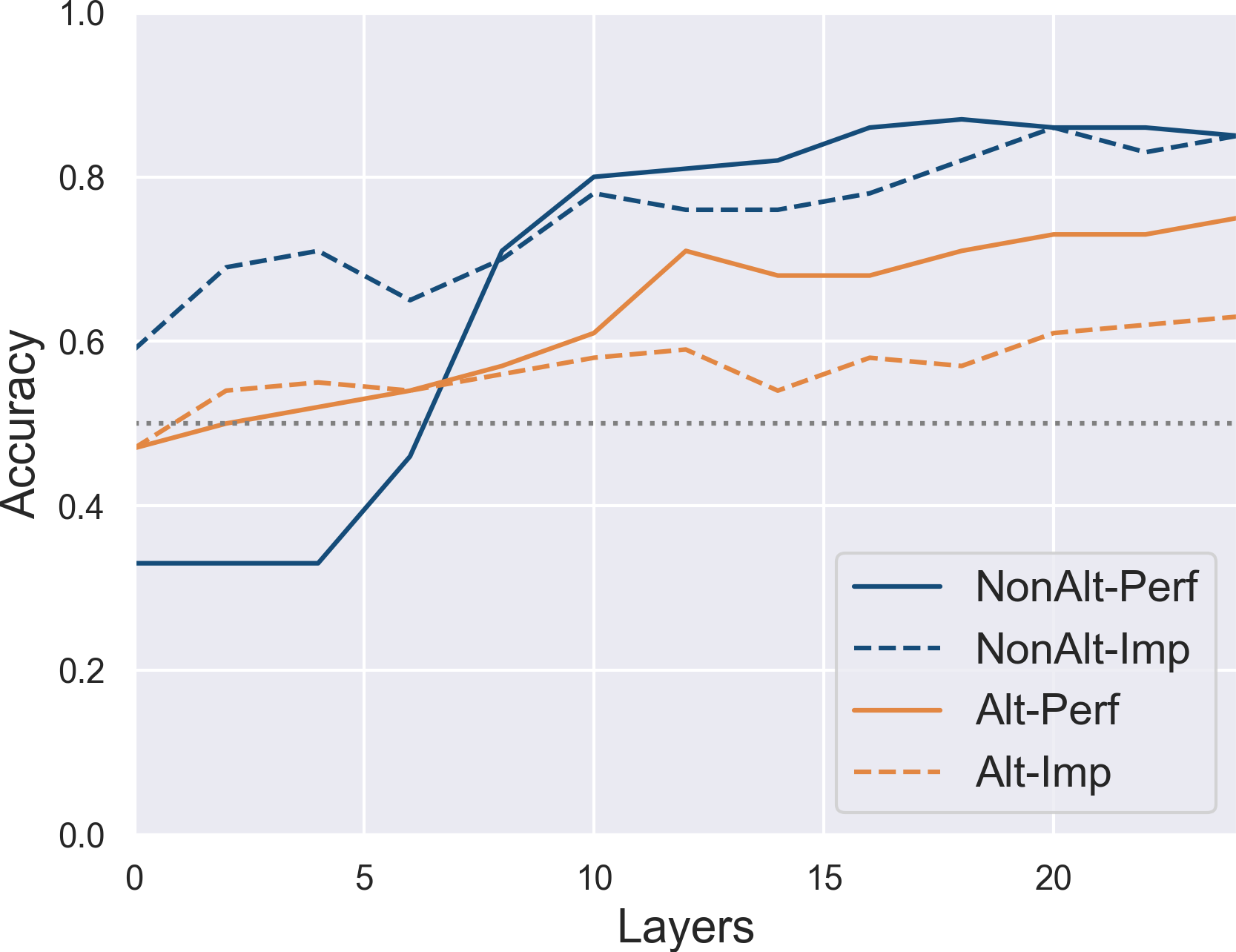}
\end{minipage}\hfill
\begin{minipage}{0.5\linewidth}
\centering
  \includegraphics[scale=0.48, trim=0mm 0mm 0mm 0mm ]{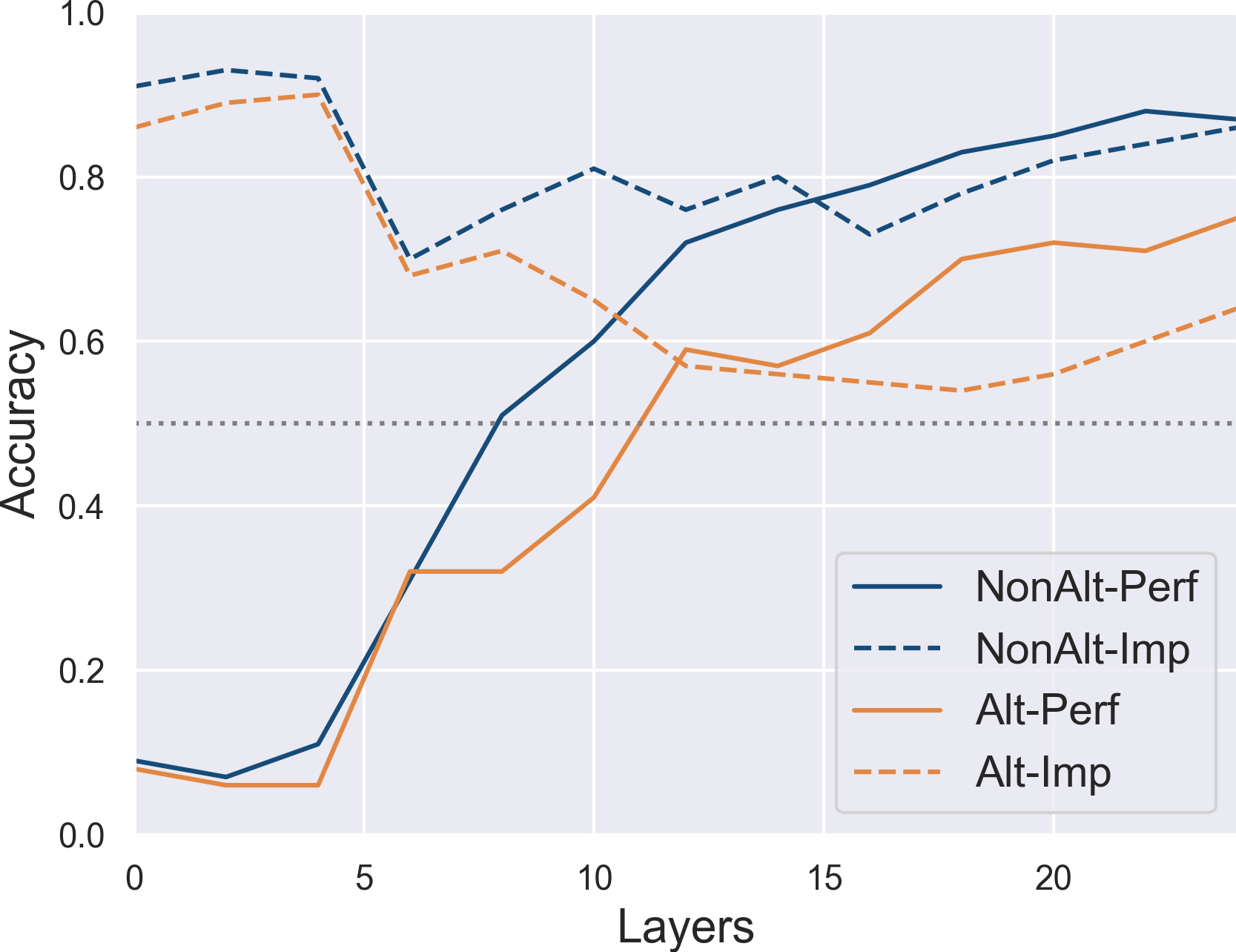}
\end{minipage}
\caption{Performance of BERT-large on iterative masking (left) and aspect inference (right) for target verbs.
\textit{Perf} and \textit{Imp} denote perfective and imperfective aspect in non-alternative (\textit{NonAlt}) and alternative (\textit{Alt}) contexts.
Black dotted lines indicate random guessing between perfective and imperfective.}
\label{types1}
\end{figure*}

\subsection{Behavioral Probing}
\label{behavioral}

First, we probe BERT and RoBERTa as Masked Language Models (MLM), in the alternative vs.~non-alternative contexts.  We evaluate the model's ability to predict aspect in the context by measuring its preference for particular grammatical forms. Typically in this task, the model is prompted to fill in the MASK given the context. The model is deemed successful if it assigns a higher probability to the correct form~\cite{marvin-linzen-2018-targeted,
lasri-etal-2022-bert,10.1162/tacl_a_00554}.

Since Russian is a morphologically rich language, with heavy inflection, its words are often split into segments during tokenization.  This is especially relevant for verbal forms, which have multiple inflectional and derivational affixes. Considering this challenge, and the fact that aspect can be marked in the prefix, the stem, or the suffix of a verb, we performed and compared two types of behavioral probing: iterative masking and aspect inference.

\VSPACEPAR{Iterative masking} entails several {\em iterations} of filling the mask.\footnote{For full detail, please see the algorithm in Appendix~\ref{alg:iterative}.} 
\comment{??? position t never changes and is never referenced.  so I removed it -- too much notation.  --- First, we pre-segment the verb in the target position $t$ into a list of $n$ sub-tokens $x^t = [x^t_1 ... x^t_n]$, where $n \geq 1$.
Then we feed the input sequence~\textbf{$\mathbf{X}$} to the model $n$ times.  On the first iteration, we replace all $n$ target tokens of $x^t$ with one [MASK]\comment{??? does anyone care about this tiny detail? --- footnote: Alternatively for RoBERTa, <mask>.} token to get $P(x^t_1|\mathbf{X}\setminus x^t)$.\  
On each $i$-th iteration, $i > 1$, we feed \textbf{$\mathbf{X}$} as input, with  the tokens up to $i$, $x^t_1 ... x^t_{i-1}$ unmasked, and replace all remaining tokens $x^t_i... x^t_n$ with one [MASK].  We accumulate the probabilities $P(x^t_i| X\setminus x^t, x^t_1...x^t_{i-1})$ that the model assigns to each target token $x^t_i$. 
After the final $n$-th iteration, we calculate target probability as the {\em average} of the accumulated conditional probabilities: $ P(x^t) \coloneqq \frac{1}{n}\sum_{i=1}^{n} P(x^t_i) $.  We perform iterative masking for each instance twice: for the perfective and imperfective forms---and compare two target probabilities.  
}
First, we pre-segment the target verb $V$ in the input sequence $\mathbf{X}$ into a list of $n$ sub-word tokens $V = [V_1 ... V_n]$, where $n \geq 1$.
Then we feed the input sequence~\textbf{$\mathbf{X}$} to the model $n$ times.  On the first iteration, we replace all $n$ target tokens of $V$ with one [MASK] token to get $P(V_1|\mathbf{X}\setminus V)$.\  
On each $i$-th iteration, $i > 1$, we feed \textbf{$\mathbf{X}$} as input, with  the tokens up to $i$, $V_1 ... V_{i-1}$ unmasked, and replace all remaining tokens $V_i... V_n$ with one [MASK].  We accumulate the probabilities $P(V_i| X\setminus V, \ V_1...V_{i-1})$ that the model assigns to each target token $V_i$. 
After the final $n$-th iteration, we calculate target verb's probability as the {\em average} of the accumulated conditional probabilities: $ P(V) \coloneqq \frac{1}{n}\sum_{i=1}^{n} P(V_i) $.  We perform iterative masking for each instance twice: for the perfective and imperfective forms---and compare the two target probabilities.  

We evaluate the model's performance by dropping one layer at a time. 
For BERT-large, Figure~\ref{types1} (left) shows consistently higher performance for both aspect forms in non-alternative contexts on layers deeper than layer 15, with peak performance (85-88\% accuracy) achieved in the final 8 layers of the model.
BERT-base yields analogous results, although performance is slightly lower in the final layers, and its ability to predict both aspects steadily improves after layer 6 (Figure~\ref{fig:types-base}, Appendix~\ref{sec:masking}). 

Performance on alternative contexts is significantly lower---since both aspect forms fit the context, the LMs show less preference for either aspect. 
Although we expect accuracy to be $\approx 50$\% in alternative contexts, BERT picks the expected form more often.
This may indicate the tendency of LMs to be more conservative when judging grammaticality~\cite{prange-wong-2023-reanalyzing}.\comment{??? how is grammaticality helping?}
However, the probabilities assigned to the expected and complementary forms in alternative contexts are much closer together than 
in non-alternative contexts, particularly after layer 15 (see Figure~\ref{fig:diff} in Appendix~\ref{sec:masking}). 

For RoBERTa-large, iterative masking shows significantly lower performance across all layers, the ability to differentiate between aspects is observed only after layer 18 in non-alternative contexts, see Figure~\ref{fig:types-roberta} in Appendix~\ref{sec:masking}. 

\VSPACEPAR{Aspect inference} is a method based on verbs in the model's dictionary, which are {\em not} segmented into sub-words---call these {\em complete} verb forms.  We feed the input sequence~\textbf{$\mathbf{X}$} to the model only once,
\comment{??? i thought we just said NO segmenting? ---replacing all $n$ segments of the target verb with a [MASK] token, where $n \geq 1$.}
replacing the target verb with a [MASK] token.  We gather the top-$k$ most probable tokens for the [MASK] position, and for each token we check whether it is also a complete verb form, with a known aspect. Then, we calculate aspect \textit{preference}: e.g., preference for perfective aspect is given by: 
$$P(\textit{perf}) = \sum_{i=1}^{k} {1}_{\{\exists \; \text{aspect = perf}\}} \cdot P(x_{i})$$

$P(x_{i})$ is the probability assigned by the model to a complete verb $x_i$.  The parameter $k$ is set to 10\% (12K tokens) of the model's vocabulary.  If most forms are perfective and have higher probabilities, we conclude that the model \textit{systematically} prefers perfective in the target position.  

As Figure~\ref{types1} (right) shows, performance of BERT-large improves steadily for both aspects after layer 15 (after 8 for BERT-base).
In the last 6 layers, aspect inference shows a similar performance to iterative masking (82-88\% accuracy in non-alternative contexts).
Our observations suggest that the capability to differentiate aspects develops after layers 12--14 for BERT-large (6--8 for BERT-base).   

Setting $k$ to 1\% of the vocabulary size gives a similar performance, except for the first 2 layers (
Figure~\ref{fig:large-1} in Appendix~\ref{sec:masking}).  From layer 0 to layer 12 for BERT-large (0--8 for BERT-base), the model seems to favor one aspect over the other.
To investigate 
this tendency, and whether predictions in early layers are conclusive, we inspected how many words out of the top-$k$ for the masked position are complete verbs, for $k=$1.2K and $k=$12K. See further details in Appendix~\ref{sec:masking}. 

\comment{??? is roberta important here?  should it just be in appendix?  :/ ...  no figures in the main paper. Answer: I will keep it.}
For RoBERTa-large, the pattern of performance on aspect inference is similar to BERT-base, although in the early and middle layers the model seems to favor perfective first and then imperfective, unlike BERT.  On layers 15--20, RoBERTa starts to differentiate between aspects.  The last 5 layers perform similarly to the last 5 layers of BERT-base and BERT-large.
As was noted by~\citet{belinkov-2022-probing}, the results of probing depend on the probed model, its original task, and the pre-training dataset.  A much smaller vocabulary (50K for RoBERTa vs. 120K for BERT) and a different pre-training dataset can cause differences between the probed models, which requires further studies.

Aspect inference is similar to syntactic evaluations by~\citet{newman-etal-2021-refining}. These evaluations address the model's \textit{systematicity} by conjugating a large set of verbs\footnote{The authors consider only {\em unsegmented} verb forms present in the models' vocabulary.} and checking the model's \textit{likely behavior}, by computing the probability that the models place on the correct form given the context.  To get a higher score, the model must conjugate more verbs correctly, instead of only preferring some well-conjugated form.  The authors show that neural models 
prefer to correctly conjugate verbs they deem likely in the target position.



\begin{figure}[t]
\center
  \includegraphics[scale=0.27]{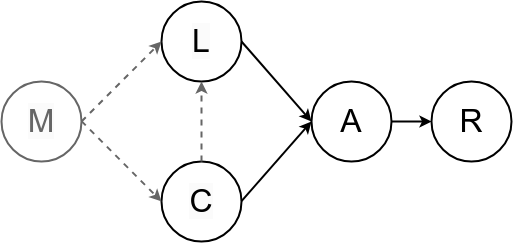}
  \caption{Causal model of dependencies between intended meaning (M) of instance, lemma of target verb (L), context (C), choice of aspect  (A), and contextual representation (R) of target verb.}
  \label{fig:model}
\end{figure}

\subsection{Causal Probing}
\label{sec:causal-probing}

In the following experiments on causal probing, we continue with {\em aspect inference} to estimate the model's behavior, and with BERT-large, due to its superior performance in the final layers. 
We use all layers for causal probing.\comment{??? therefore, }
Although aspect inference is limited to verbs that are {\em complete} (unsegmented), this method gives a reliable assessment, since the percentage of complete verbs among the top-$k$ predictions is high.
Further, aspect prediction does not depend on the lemma of the verb, on its original form in the instance and its aspect pair, or on segmentation.
It is also significantly faster than iterative masking.

We use a causal model of relations between the choice of aspect $A$ and the intended meaning $M$ conveyed in the context: $M$ affects the choice of lemma $L$ for the target verb and the choice of the surrounding context $C$ (Figure~\ref{fig:model}).  Since aspect is a grammatical category, we do not draw a direct connection between $M$ and $A$. We focus on interventions into $L$ and $C$: we will (1) remove the effect of lemmas by masking them and (2) alter the semantics of the context $C$ by replacing the model's original representation with a counterfactual one.
To generate counterfactual representations, we use the {\em AlterRep} method~\cite{ravfogel-etal-2021-counterfactual}.  It is designed to study how the model uses a particular linguistic feature---by altering the representation\comment{??? representation or VALUE? Answer: we cannot alter the value.} of the studied feature, and investigating whether the resulting changes in the model's behavior agree with linguistic theory. 

\VSPACEPAR{Boundedness:} Aspect differs from previously studied syntactic phenomena---such as number agreement between subject and verb, etc.---for which it is easy to identify linguistic features that are directly involved in the phenomenon and can be used for causal probing.  To probe aspect, we leverage the semantics of the context: in particular, how the meaning of \textit{bounded vs. unbounded action} affects the choice of aspect.  In Example (1), e.g., the imperfective verb form and the adverb \textcyr{``всегда''} (\textit{always}) conveyed that action is unbounded.

We identify \CUEW{cue words} in the context---``Resultative'', ``Inception'' words, etc.---which give \CUEW{cues} regarding the boundedness of the action described by the verb and determine the choice of aspect.\footnote{The complete list of cue words is given in Appendix~\ref{sec:dataset}.}
The action is bounded if \underline{the target verb}:
\begin{enumeratorCompact}
    \item has a ``Resultative'' adverbial modifier or argument, e.g.:
    \textcyr{``\CUEW{Внезапно} она все \underline{поняла}.''}\\ (\textit{\CUEW{Suddenly}, she \underline{understood} everything.}) 

    \item has a ``Duration'' argument, e.g.,\\
    \textcyr{``Она \underline{пробежала} круг \CUEW{за 5 минут}.''}\\
    (\textit{She \underline{ran} the lap \CUEW{in 5 minutes}.})
    
    \item is a complement of a ``Capability'' verb, e.g.:\\
    \textcyr{(``Она \CUEW{смогла} его \underline{понять}.'')\\
    (\textit{She \CUEW{was able to} \underline{understand} him.})}
    
    \item is a complement of a ``Forget'' verb, e.g.:\\
    \textcyr{``Она \CUEW{забыла} \underline{зарядить} телефон.''}\\
    (\textit{She \CUEW{forgot} \underline{to charge} the phone)}
\end{enumeratorCompact}

\noindent
The instance is unbounded if the target verb:
\begin{enumeratorCompact}

\item is a complement of an ``Inception'' verb, e.g.:\\
\textcyr{``Он \CUEW{начал} \underline{петь}.''} (\textit{He \CUEW{began} \underline{to sing}}). 

\item has an ``Iterative'' adverbial modifier, e.g.,\\
\textcyr{``\underline{Гулять} в лесу \CUEW{каждый вечер}.''} \\(\textit{\underline{To walk} in the forest \CUEW{every evening}.})

\item is a complement of a ``Like'' verb, e.g.:\\
\textcyr{``Она \CUEW{любила} \underline{читать}.''} (\textit{She \CUEW{liked} \underline{to read}.)}

\end{enumeratorCompact}


\VSPACEPAR{INLP:} Following~\citet{ravfogel-etal-2021-counterfactual}, we denote by $T$ a set of words in context, $H$ the set of contextual representations of $T$, $\vec{h_t} \in$ $\mathbb{R}^d$ the representation of word $t$. 
Let $F$ be a linguistic feature encoded in $H$---here: {\em boundedness}.  INLP defines the ``feature sub-space'' $R$ of the original representation space where $F$ is encoded.  $R$ is spanned by $m$ learned directions\comment{??? we never define what m is ... :/ is it the number of layers in BERT?  unclear.. Answer: m is a parameter, number of classifiers}---weight vectors of $m$ linear classifiers trained to predict $F$ given $H$, where all $m$ are mutually orthogonal.  
Counterfactual representations $\vec{h^+_t}$ and $\vec{h^-_t}$ encode that the word $t$ has positive or negative values of $F$, regardless of the true value of $F$ encoded in the original $\vec{h_t}$.  Counterfactuals are generated by pushing $\vec{h_t}$ further away in the opposite directions from the separating planes learned by $m$ classifiers, see Appendix~\ref{app:inlp}.

We use {\em boundedness} as the feature $F$ encoded in the representations $H$.  $F$ has two values: '${+}$' if the action in the context is bounded, or '${-}$' if unbounded. We train $m$ SVM classifiers to define a sub-space of boundedness $R$ and use it to manipulate the value of $F$ in the target verb by generating counterfactual context representations $\vec{h^+_t}$ and $\vec{h^-_t}$.

\begin{figure}
\center
  \includegraphics[scale=0.5]{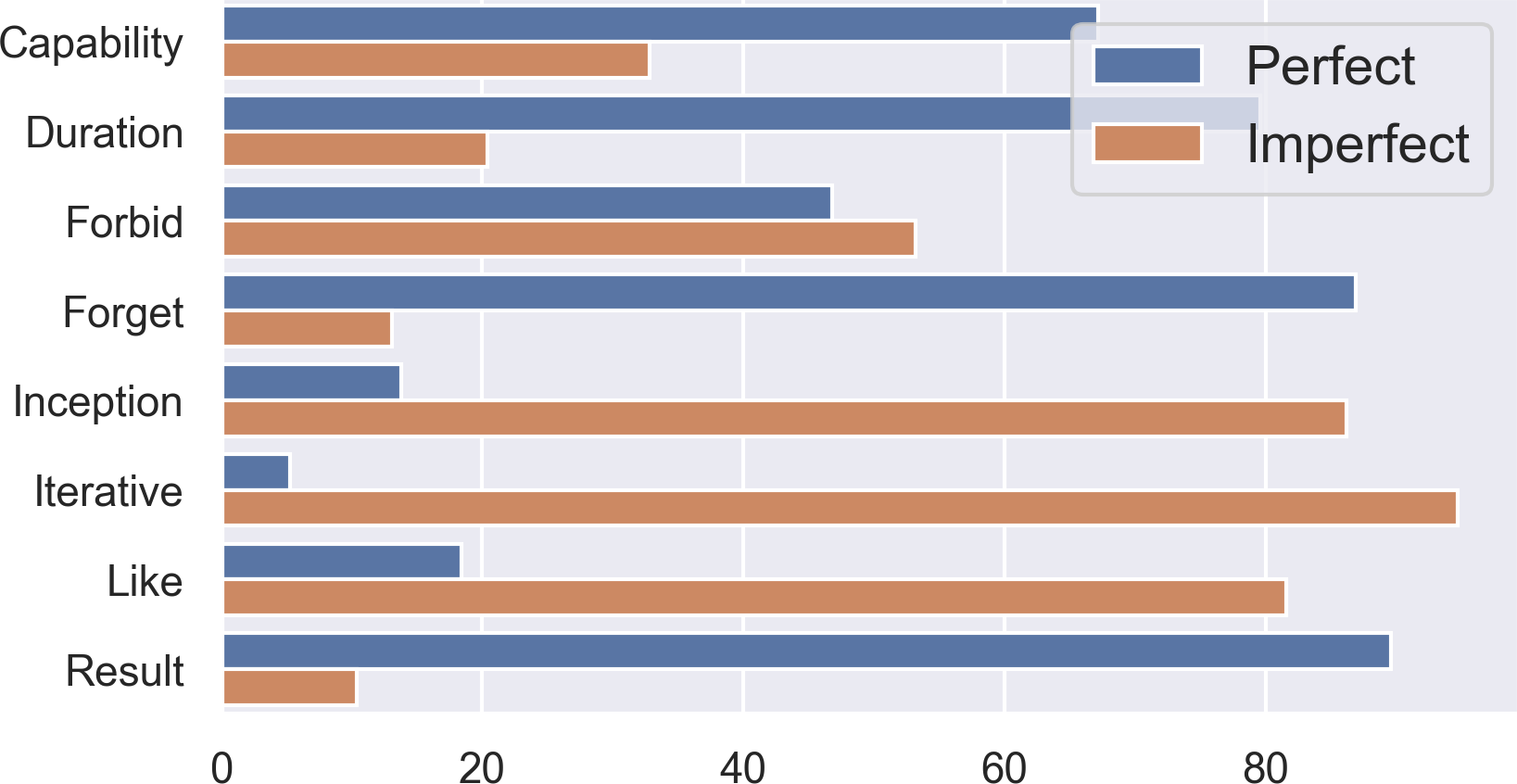}
  \caption{Percentage of sentences with detected cue words where target verb is perfective or imperfective.}
  \label{fig:sem_type} 
\end{figure}

\begin{figure*}[t!]
\begin{minipage}{0.5\linewidth}
\centering
  \includegraphics[scale=0.48, trim=0mm 0mm 0mm 0 ]{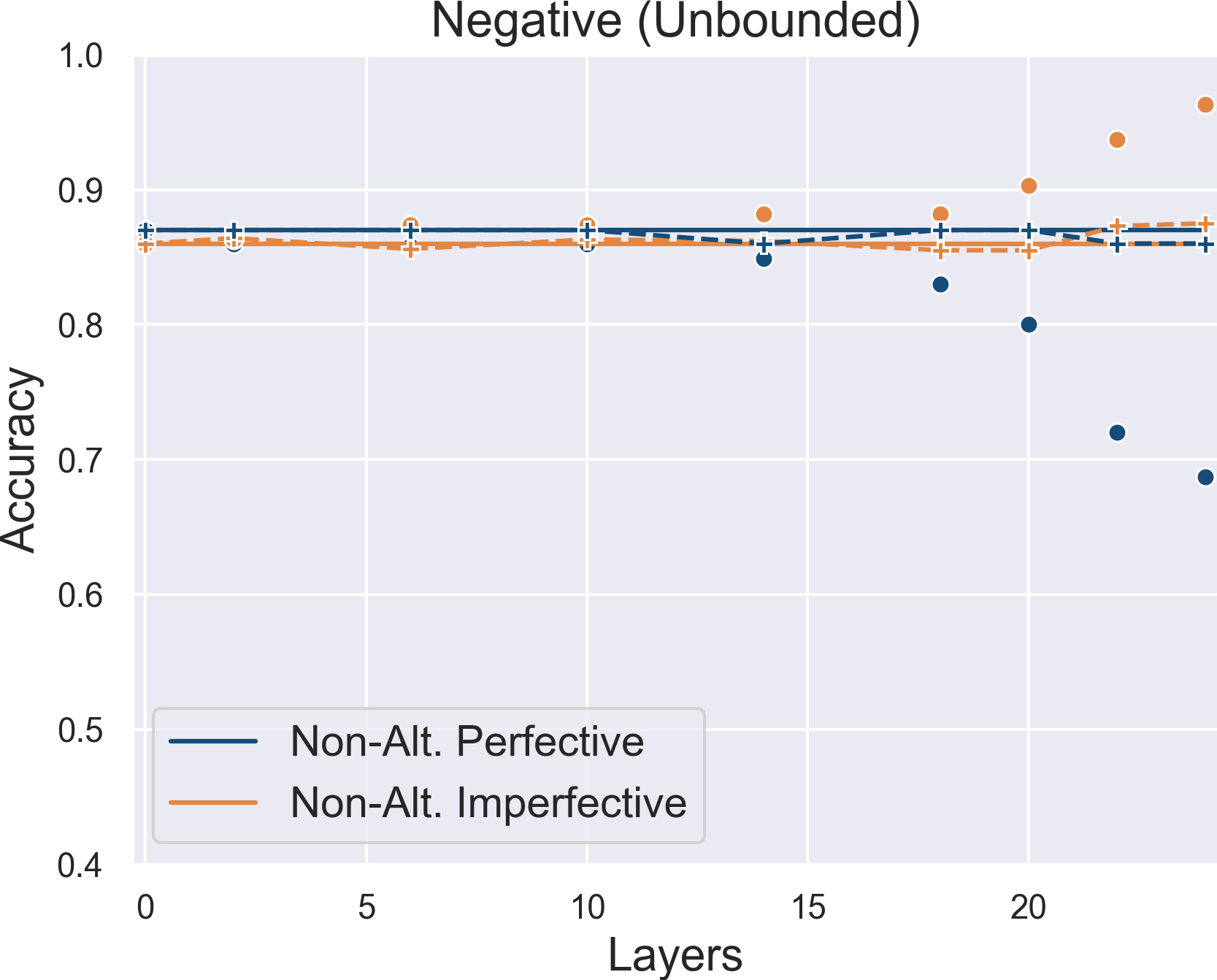}
\end{minipage}\hfill
\begin{minipage}{0.5\linewidth}
\centering
  \includegraphics[scale=0.48, trim=0mm 0mm 0mm 0 ]{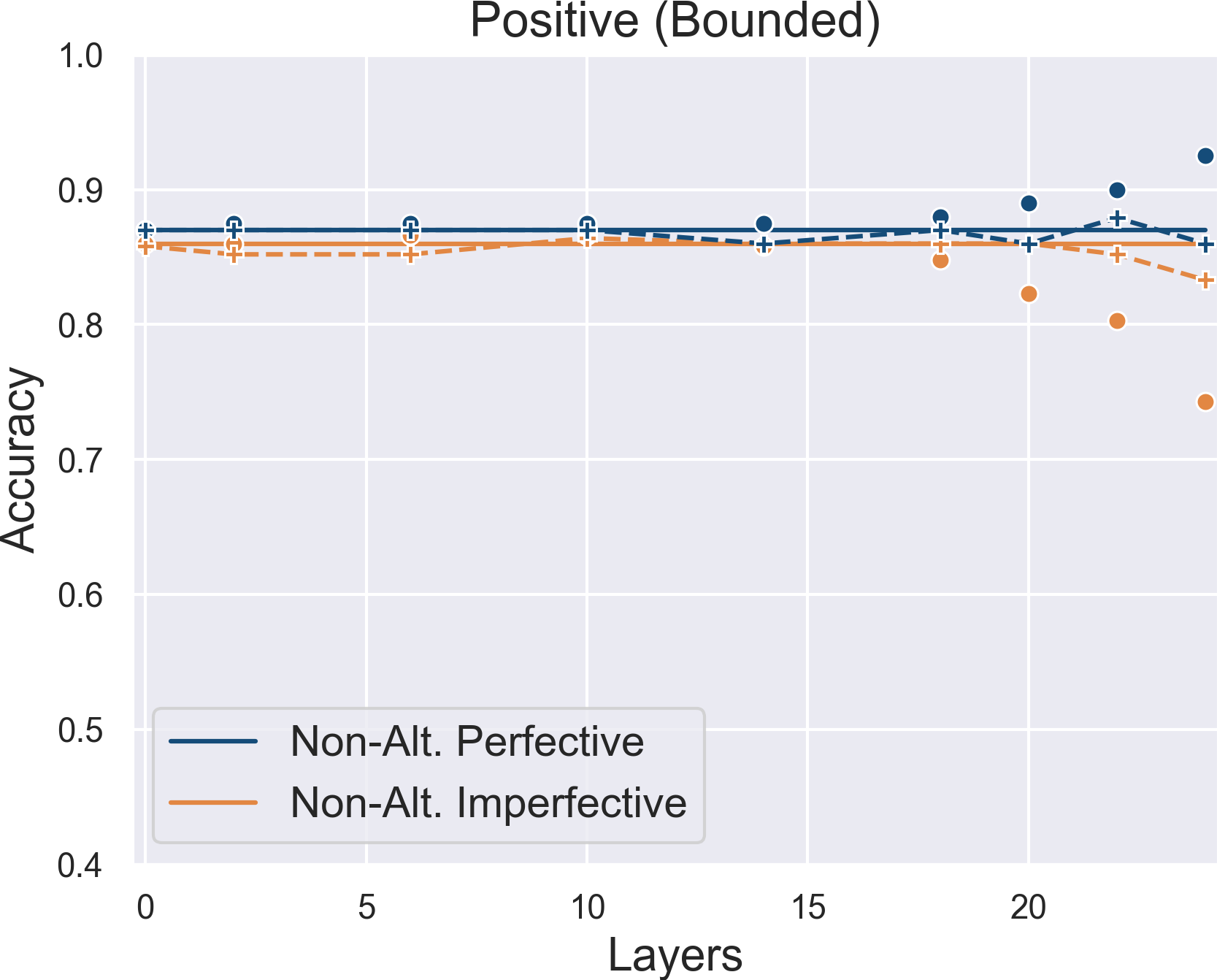}
\end{minipage}
\begin{minipage}{0.5\linewidth}
\centering
  \includegraphics[scale=0.48, trim=0mm 0mm 0mm 0 ]{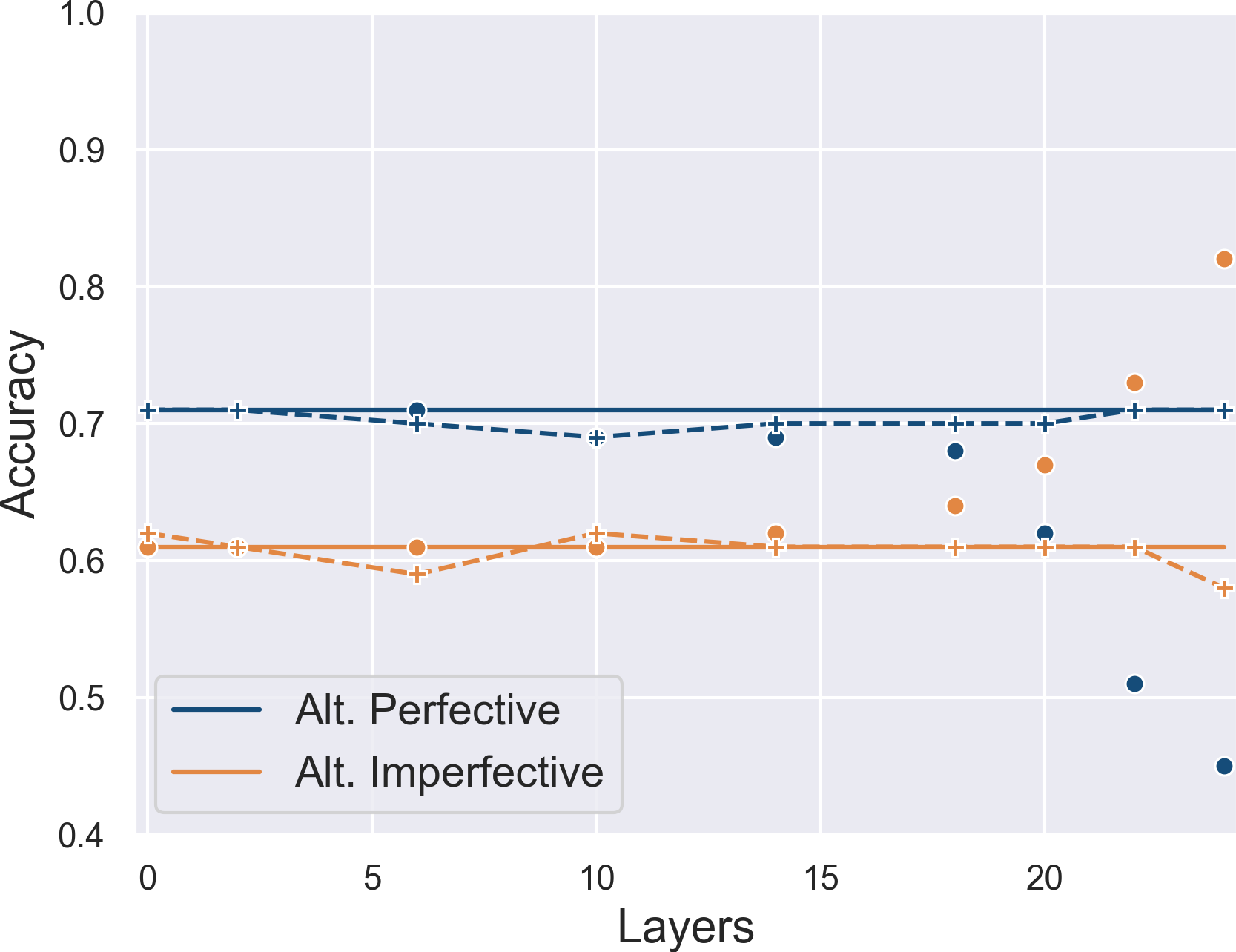}
\end{minipage}\hfill
\begin{minipage}{0.5\linewidth}
\centering
  \includegraphics[scale=0.48, trim=0mm 0mm 0mm 0 ]{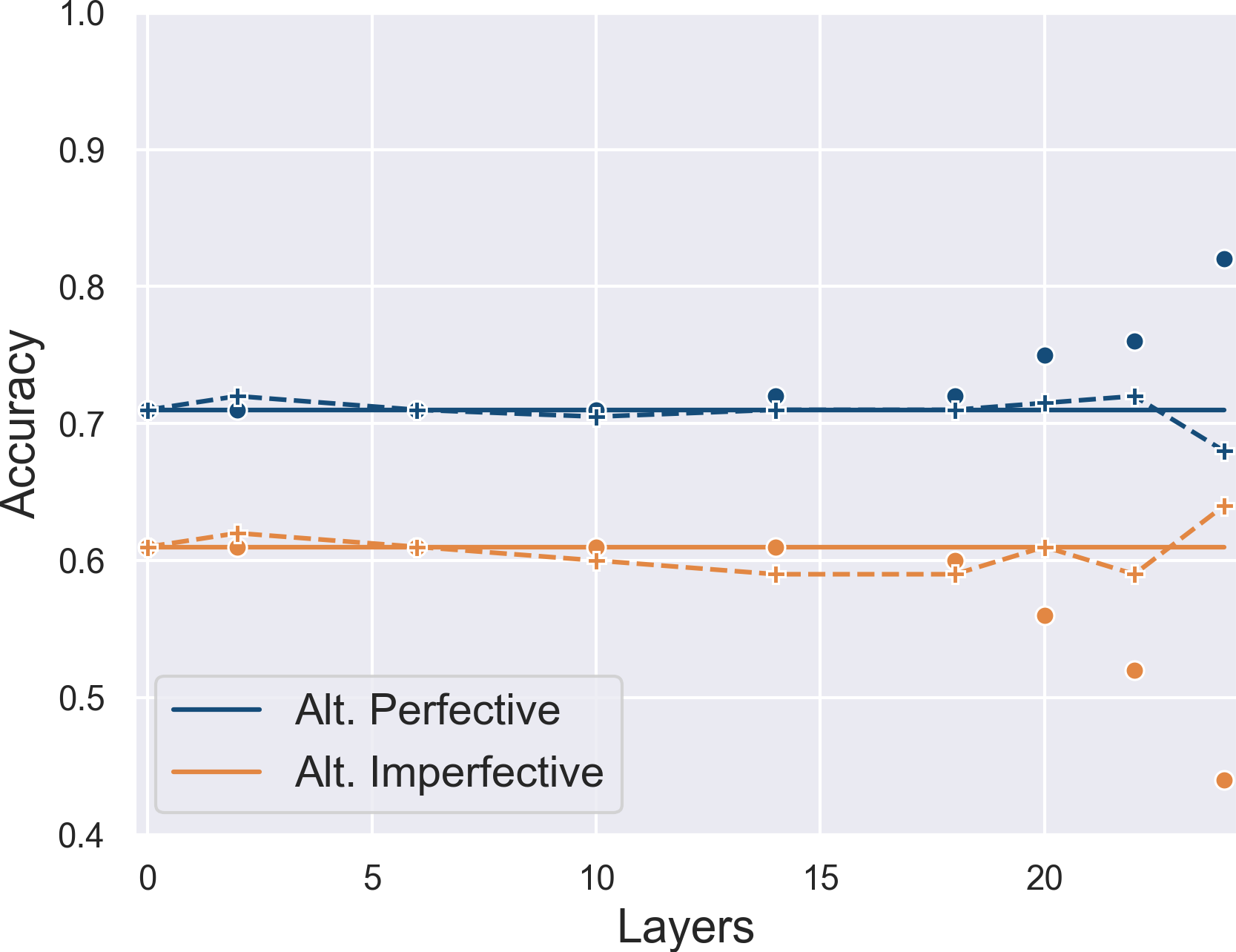}
\end{minipage}
\caption{Accuracy of predicting correct (expected) aspect, using {\em aspect inference} method after intervention on BERT-large representations. Top plots---non-alternative contexts; bottom plots---alternative contexts. 
Left plots---negative intervention: toward unbounded action.  Right plots---positive intervention: toward bounded action.
\textbf{Flat} lines show performance before intervention; \textbf{dots}---after intervention. \textbf{Dashed} lines---after random interventions.} 
\label{inlp}
\end{figure*}

\VSPACEPAR{Data:} To train INLP classifiers, we need a dataset with pairs of contexts where the described action is bounded or unbounded.  We collect instances automatically from the Omnia corpus and make sure that they do not appear in the test data.
In a sentence parsed with a dependency parser~\cite{burtsev-etal-2018-deeppavlov}, we pick the verb as a target only if it participates in syntactic relations with one or more cue words indicating boundedness.

We collect 8160 instances for each value of $F$. The choice of the types of relations was guided by grammatical rules, materials for language teaching~\cite{kagan2014russian,volkovarussian}, and statistics derived from the SynTagRus corpus~\cite{droganova2018data}, see Figure~\ref{fig:sem_type}. 
In some constructions, both perfective and imperfective verb forms can be found.\footnote{E.g., we can find examples of perfective verbs that depend on ``Inception'' verbs, but they are very infrequent in the corpus.

More complex is the situation with certain {\em ambiguous} cue words, e.g., \textcyr{\CUEW{нельзя}} ({\em impossible/forbidden}), which can appear equally with either aspect:
its complement can be imperfective---\textcyr{``Здесь \CUEW{нельзя} \underline{курить}'',  (\textit{\underline{Smoking} is \CUEW{prohibited} here)}}, 
or perfective---\textcyr{``\CUEW{Нельзя} \underline{закурить} при сильном ветре'',  (\textit{\CUEW{Impossible} to \underline{smoke} in  strong wind)}}.

We exclude instances of such constructions from the data.
}

\VSPACEPAR{Training INLP:} For every collected instance, we replace the target verb with a [MASK] token---to remove the influence from its lemma---and feed INLP classifiers with \textit{contextual vectors of the cue words} from different BERT layers. 
Since the cue may be segmented into multiple sub-word tokens, we average the representation from the vectors of all cue segments.  See training details in Appendix~\ref{app:inlp}. 






\VSPACEPAR{Effect of Interventions:}
To assess the impact of counterfactuals, we measure the accuracy of aspect prediction using the aspect inference method.
As in~\citet{ravfogel-etal-2021-counterfactual}, for each sentence, we mask the target verb, start the forward pass, perform interventions on the verb representation at the specific layer, and continue the forward pass.  Then, we retrieve the top-$k$ tokens for the masked position and compute the model's preference for aspect.  Figure~\ref{inlp} shows the results on the data used for behavioral probing, for non-alternative (top plots) vs.~alternative contexts (bottom).  The left plots display the results using negative counterfactuals---shifting representations toward unbounded action, and the right plots---positive counterfactuals, shifting toward bounded action.  
The X-axis indicates the layer at which the intervention is performed.

The most significant changes in the accuracy of predicting aspect in the masked position are seen in the model's latter layers (post layer 20)---compare the flat lines, indicating performance before interventions vs.~dots, indicating an intervention.
This trend is observed for both aspects, using negative and positive interventions in the alternative and non-alternative contexts.
It agrees with the findings of behavioral probing, where the peak performance for both aspects was evident mostly in the final layers. The results align with our hypothesis and grammatical theory: shifting representations toward the ``unbounded'' sub-space improves the predictions of imperfective aspect and significantly increases the error in predictions of perfective aspect; moving representations in the opposite direction of the ``bounded'' sub-space has the opposite effect---the accuracy of perfective rises, while the accuracy of imperfective deteriorates.

Negative interventions influence imperfective in both alternative and non-alternative contexts: the maximum accuracy shift is $+$21\% and $+$10.3\%, respectively, in layer 24.  Similarly for positive interventions: the maximum accuracy shift for imperfective is $-$17\% in alternative and $-$11.7\% in non-alternative contexts. The impact of negative interventions on perfective is higher for alternative ($-$26\%) and non-alternative contexts ($-$18.3\%), as compared to the effect of positive interventions: in alternative contexts, perfective accuracy increases by 11\%, and for non-alternative---by 5.5\%.
The plots show that interventions have stronger impacts in alternative contexts. 
Further, negative intervention has a stronger effect in both types of contexts.\footnote{We observe a similar patterns for BERT-base, see Figure~\ref{types-base} in Appendix~\ref{app:base}.}
This could be caused by the data used to train the INLP classifiers---cue words indicating unbounded action appear with imperfective verbs more consistently, see Figure~\ref{fig:sem_type}.

\begin{figure*}[t!]
\center
  \includegraphics[scale=0.415]{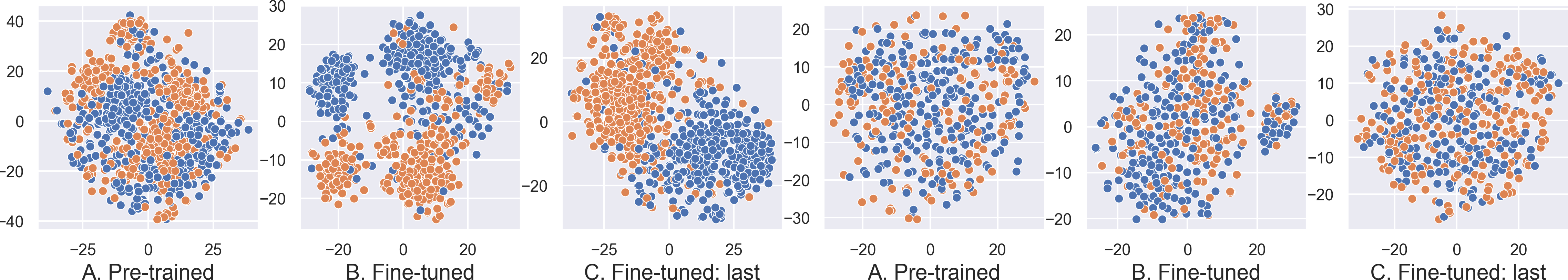}
  \caption{t-SNE visualization of representations from BERT-large layer 24 for masked target verbs in {\em non-alternative} ({\bf left} 3 plots) and {\em alternative} ({\bf right} 3 plots) contexts using A.~pretrained models, B.~fine-tuned models, and C.~models with fine-tuned last layers. 
  Orange indicates imperfective, and blue---perfective.} 
  \label{fig:emb}
\end{figure*}

We apply causal probing to RoBERTa-large and observe a similar pattern: only layers 18-24 are affected by interventions (Figure~\ref{fig:inlp-roberta} in Appendix~\ref{app:base}).
The influence of intervention\comment{on predicting aspect} is the same as for BERT.  However, the difference between accuracy shift in alternative vs.~non-alternative contexts is not as striking as for BERT\comment{-large ??? we already said ONLY large ...}.
\comment{??? is this important to note here?  seems minor --> }
\comment{The maximum change in accuracy is seen in layer 22, unlike for BERT, where the maximum shift is in layer 24.}

\noindent\textbf{Selectivity:} To ensure the \textit{selectivity} of the probe, we verify that random changes in representations do not impact aspect prediction in the same manner. Random counterfactuals were generated using 20 random sub-spaces.  Dashed lines in Figure~\ref{inlp} show that changes in accuracy are smaller and do not follow the pattern observed with altering boundedness.  Additionally, we ensure that the interventions targeting context semantics do not affect the predictions of other grammatical categories in the same way as they affect aspect.  
We perform the same experiment, but measure the accuracy of predicting the grammatical number of the masked target verbs.  We choose number because it has no relation to aspect and frequently appears in verb forms.
The results indicate no significant change in the prediction of number on any layer (Figure~\ref{fig:number}, Appendix~\ref{sec:number}).

\VSPACEPAR{Probing with Iterative Masking:}  We checked whether causal probing shows similar results with different methods of evaluating the model's performance.
Using iterative masking instead of aspect inference confirms the above observations.  The main difference is the absolute value of the accuracy shift: it is in the range 2\%--18\% for the last layers of BERT-large.

\subsection{Fine-tuning for Aspect Prediction}

\begin{table}[t]
  \begin{center}
      \scalebox{0.81}{
  \begin{tabular}{l|rr|rr}
\hline
    \multicolumn{1}{l|}{} & \multicolumn{2}{c|}{non-alternative}  & \multicolumn{2}{c}{alternative} \\
      Model & $F_{0.5}^{\text{perf}}$  & $F_{0.5}^{\text{imp}}$ & $F_{0.5}^{\text{perf}}$  & $F_{0.5}^{\text{imp}}$ \\\hline
    A. Pretrained               & 36.3       & 49.2 	  & 54.0 & 51.1  \\
    B. Fine-tuned               & 85.9       & 84.0       & 67.5 & 57.0  \\
    C. Fine-tuned last 5 layers & {\bf 88.5} & {\bf 88.0} & 69.1 & 64.2  \\
\hline
  \end{tabular}
  }
  \caption{Performance in terms of $F_{0.5}$ for aspect prediction in non-alternative and alternative contexts.}
  \label{table:perf}
  \end{center}
\end{table}

To utilize the information found through probing, we fine-tune BERT-large for the aspect prediction task.  We formulate the task as a 2-way classification, where the model predicts whether the masked verb is perfective or imperfective.  We use the SynTagRus corpus to create training and validation data.\footnote{The model is trained only on verb forms that have aspect tags in their morphological analysis.} Inspired by the probing results, we fine-tune layers 20-24 of the BERT encoder and the last classification layer, keeping all other layers frozen.

Table~\ref{table:perf} shows the classification performance in three experiments: A.~prior to fine-tuning; B.~after fine-tuning all layers; and C.~after fine-tuning the final 5 layers.
Rows B--C show performance averaged across 5 fine-tuning runs. Freezing layers up to layer 20 speeds up fine-tuning and increases performance for aspect prediction, especially for imperfective aspect.
Fine-tuning can yield performance comparable to the performance of BERT-large as a MLM at its final layers in non-alternative contexts.
In alternative contexts, results are lower. Details on data, training, and evaluation with other layers are in Appendix~\ref{sec:finetune}.

To visualize the changes in the model's representations, 
we use t-SNE~\cite{JMLR:v9:vandermaaten08a} to project the masked verb representations onto the 2-D plane, Figure~\ref{fig:emb}.  Notably, fine-tuning the final layers results in more refined clustering of representations based on aspect.  The lack of structure in the verb representations within alternative contexts aligns with our observations from the two behavioral probing methods---consistently lower preference for either aspect form.  


\subsection{Error Analysis}
\label{sec:error-analysis}

\begin{figure}
\center
  \includegraphics[scale=0.45]{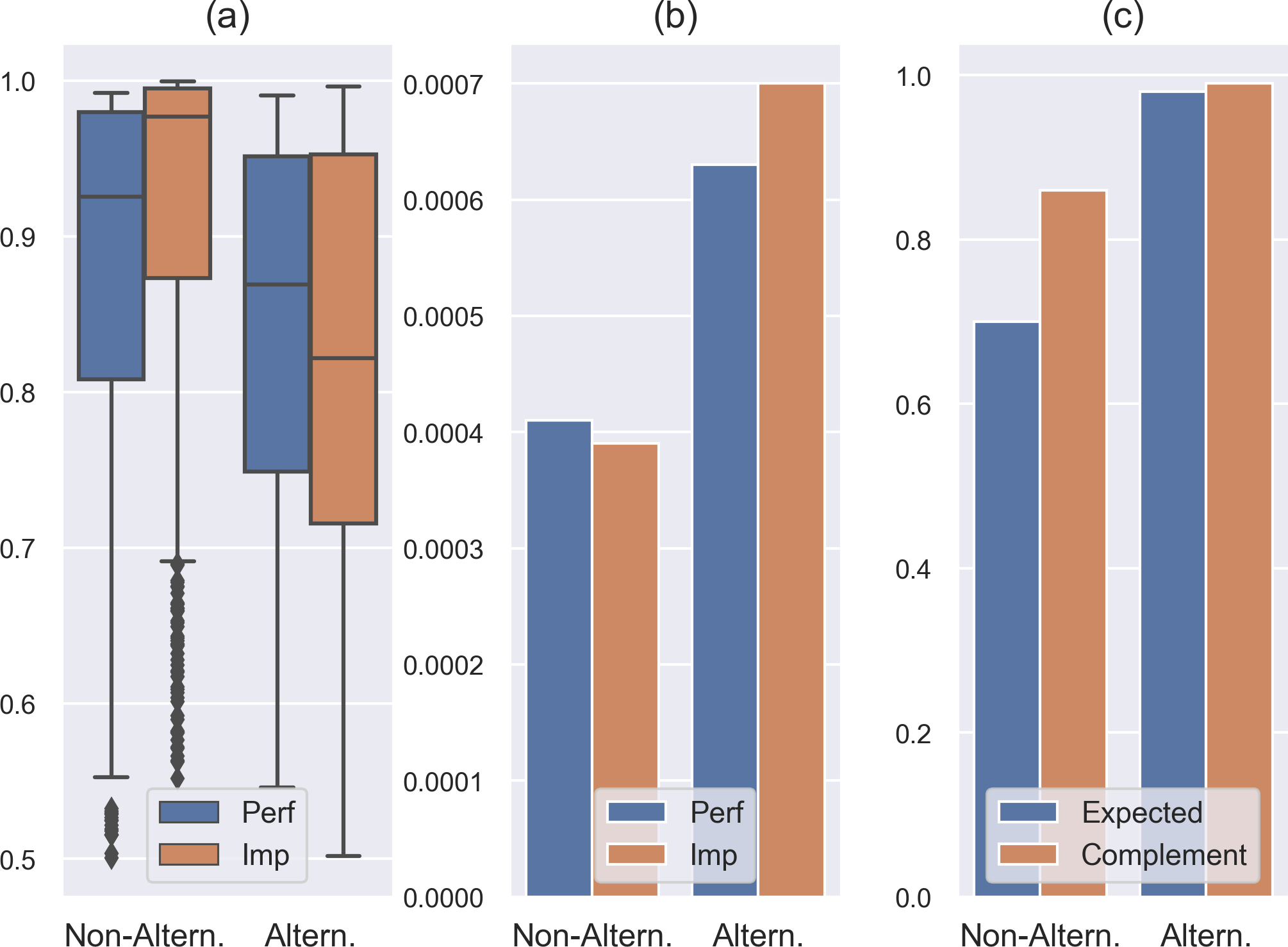}
  \caption{(a) Scores assigned to imperfective and perfective classes.
  (b) Variance of scores assigned to imperfective and perfective classes. 
  (c) Percentage of \underline{contexts lacking cue words} when the model predicts:
  \textit{Expected} aspect vs.~\textit{Complementary} aspect. }
  \label{fig:var}
\end{figure}

\noindent
{\bf Uncertainty:}  The aspect inference method does not allow us to directly calculate the {\em uncertainty} of aspect prediction for a given instance.  Therefore, we use Monte Carlo dropout~\cite{pmlr-v48-gal16} to estimate the confidence of the fine-tuned model with frozen layers.  
For every input, we repeatedly sample 20 predictions with dropout activated, and calculate the variance; see plot (b) in Figure~\ref{fig:var}.  The model has much higher predictive uncertainty for alternative contexts: BERT cannot make a preference for a particular aspect; see the orange bars indicating high variance. 

We perform an automatic and a manual analysis of both types of contexts, to examine the possible reasons why BERT struggles with the aspect, either as a MLM or fine-tuned.
We collect all contexts where BERT as MLM prefers the expected (blue bars in plot (c), Figure~\ref{fig:var}) vs.~complementary aspect (orange).  Preference is calculated using the aspect inference method.  We calculated predictive uncertainty for the preferred aspect in each of these contexts.   
Then, we manually inspected the contexts with the highest variance.  We observed the main difference---almost none of the alternative contexts contain cue words that could inform the preference for one aspect over the other.

\noindent
{\bf Absence of Cues:} Appendix \ref{sec:dataset} lists many cue words that indicate bounded vs.~unbounded action.  Although this list is not exhaustive, it provides a rough estimate of the difference between the contexts, and can aid in checking the manual analysis.  We used this list to automatically inspect how many of the contexts do not contain cue words, and to check our manual evaluation.  In Plot (c) the two left bars show that in non-alternative contexts, when BERT as a MLM predicts the complementary aspect (rather than the expected), most such contexts have no cues (orange bar, more than 85\%).
Of the contexts where the model predicted the aspect correctly, 70\% (left blue bar) also have no cues, which indicates that other types of contextual evidence must be present in the context.  This requires further study.  

Almost 100\% of alternative contexts have no cue words at all (plot (c), right two bars), which might explain why counterfactuals have more impact in alternative contexts---positive or negative interventions introduce the missing ``hints'' into the representations.  

\section{Conclusions and Future Work}


We investigate the encoding of the grammatical category of verbal aspect for Russian in PLMs---particularly, BERT and RoBERTa---via behavioral and causal probing. Encoding of aspect has not been studied to date for any language or model.  All types of probing show that these models do encode aspect and learn to distinguish between aspect forms primarily in their final layers.  Using this finding, we fine-tune BERT for aspect prediction, which leads to more effective and faster tuning.  

In line with linguistic theory, information about the {\em boundedness} of the action is encoded in the model's context representation and affects the choice of aspect: shifting representations towards the ``bounded'' space positively affects prediction of perfective forms (and negatively---of imperfective), and vice versa.
Prediction of aspect is not affected by random interventions.
We checked that the causal probe is selective and does not affect irrelevant categories, e.g., number.

A particular challenge is caused by contexts where more than one aspect form can fit grammatically and semantically, which we call {\em alternative}.  We investigated whether encodings of aspect differ in these contexts from non-alternative ones.  We find that BERT is consistently uncertain about aspect forms in alternative contexts.
Causal interventions also have a stronger effect in such contexts.  Our error analysis shows that these contexts do not have enough cues to help the model (or a human) decide which aspect to use.  

In future work, we plan to explore additional languages; investigate how transformers encode relations between verbs and the cue words; and inspect the connections between aspect and tense of verbs, and context words expressing time, by intervening in the attention weights.  We also plan to investigate aspect prediction in contexts lacking cue words, where information affecting the choice of aspect is presented in the neighboring sentences and requires reasoning.       
The practical goal is to deploy the aspect prediction in the production language teaching/learning system, to help learners master this advanced and complex feature of Russian.

\section*{Acknowledgements}

\comment{BusinessFinland grant, HIIT grant}
This research was supported in part by BusinessFinland Project ``Revita'' (Grant 42560/31/2020), 
and by a grant from the Helsinki Institute for Information Technology (HIIT).

\section*{Limitations}

This work has a number of limitations to consider: 

(A) The experimental design of the paper was limited to a single language. Aspectual systems vary significantly across languages. Therefore, adding a new language requires linguistic expertise and a new experimental setting. For Russian, we performed causal intervention in the context's meaning of boundedness and compared perfective vs. imperfective verb forms. Many languages do not have the opposition of these two forms as in Russian, and the meaning of boundedness may not be as significant for the choice of aspect in context. The closest aspect system to Russian among Slavic languages is Polish.  Probing it would require a substantial investment of resources, which our team lacks.

(B) We experimented only with masked language models available for Russian since they have access to the full context, which is more relevant for the aspect prediction task.    

(C) Due to resource constraints, we could not engage more people in data collection and annotation. While we recognize that our dataset is relatively small, we believe it is crucial to share the data we have. We hope it draws the research community's attention to the complex problem of aspect probing.

(D) We acknowledge that there is no consensus regarding several important questions among linguists studying the category of aspect in Slavic languages: the meaning of aspect opposition or whether aspect pairs represent forms of the same verb or different verbs.  There are well-founded different opinions on each of these questions.  We shape particular views for clarity of our experiments.   

(E) We also recognize that our list of cue words, which indicate the boundedness of actions, is not exhaustive.  We also ignore for now other contextual evidence indicating whether an action was completed, and whether its result is observable at the moment.  Identifying this information is more complex and, we believe, requires reasoning.  We plan to extend our work to investigate various types of contexts and larger PLMs. 

(F) Due to the page limit, we did not include the effects of the removal of the linguistic feature of boundedness in the current experiment which could be an interesting extension of the experiment in the future versions of the paper. 

(G) Our current experiments do not include an investigation of attention weights which we plan to do in future work.

\section*{Ethics Statement}

We used publicly available data, code, and models for the described experiments. 

Annotated data that we release together with this paper will be freely available for the research community to be used for extending probing experiments. Data does not have any personal information, does not identify individual people, and does not include offensive content. Annotators are volunteers who have previous experience in annotation and are aware of how the annotated data is going to be used. We also do not see any potential risk that might be caused by our work.

\bibliographystyle{acl_natbib}
\bibliography{custom}

\appendix
\section*{Appendices}

\section{Generating Aspect Test Data}
\label{sec:appendix1}

\begin{algorithm*}[t!]
\caption{Iterative Masking}
\begin{algorithmic}[1]
    \STATE \textbf{Input:} Sequence of tokens $\bm{X}$ with target verb $V$; pre-segmented target verb $V = [V_1, \dots, V_n]$ where $n \geq 1$. 
    \FOR{$i=1$ to $n$}
        \IF{$i = 1$}
            \STATE Replace all $V$ with [MASK] and feed $\bm{X}$ to BERT.
            \STATE Calculate  $P(x^1| X \setminus V)$
        \ELSE
            \STATE Keep target segments $V_1, \dots, V_{i-1}$ unmasked in $\bm{X}$.
            \STATE Replace $V_{i}...V_{n}$ with one [MASK] and feed $\bm{X}$ to BERT
            \STATE Calculate  $P(V_i|X \setminus V, V_1,\dots,V_{i-1})$
        \ENDIF
    \ENDFOR
    \STATE Get averaged probability of the target:
    \STATE $P(V) \coloneqq  \frac{1}{n}\sum_{i=1}^{n} P(V_i) $
    \STATE Execute iterative masking twice for $\bm{X}$: with perfective and imperfective target verb forms
    \STATE Compare $P(V_{\text{perf}})$ and $ P(V_{\text{imp}})$
\end{algorithmic}
\label{alg:iterative}
\end{algorithm*}

\begin{figure*}
\begin{minipage}{0.5\linewidth}
\centering
  \includegraphics[scale=0.48, trim=0mm 0mm 0mm 0 ]{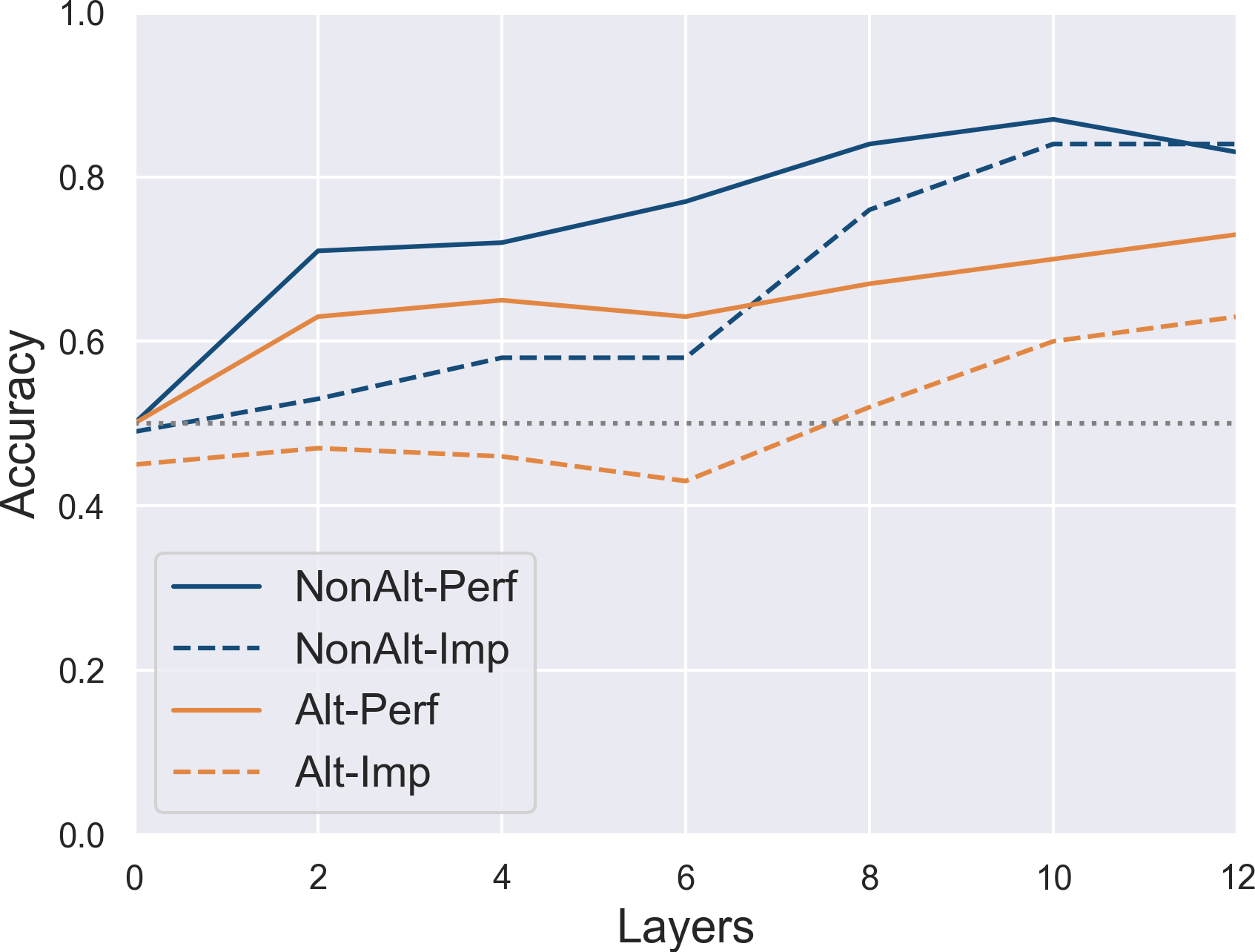}
\end{minipage}\hfill
\begin{minipage}{0.5\linewidth}
\centering
  \includegraphics[scale=0.48, trim=0mm 0mm 0mm 0 ]{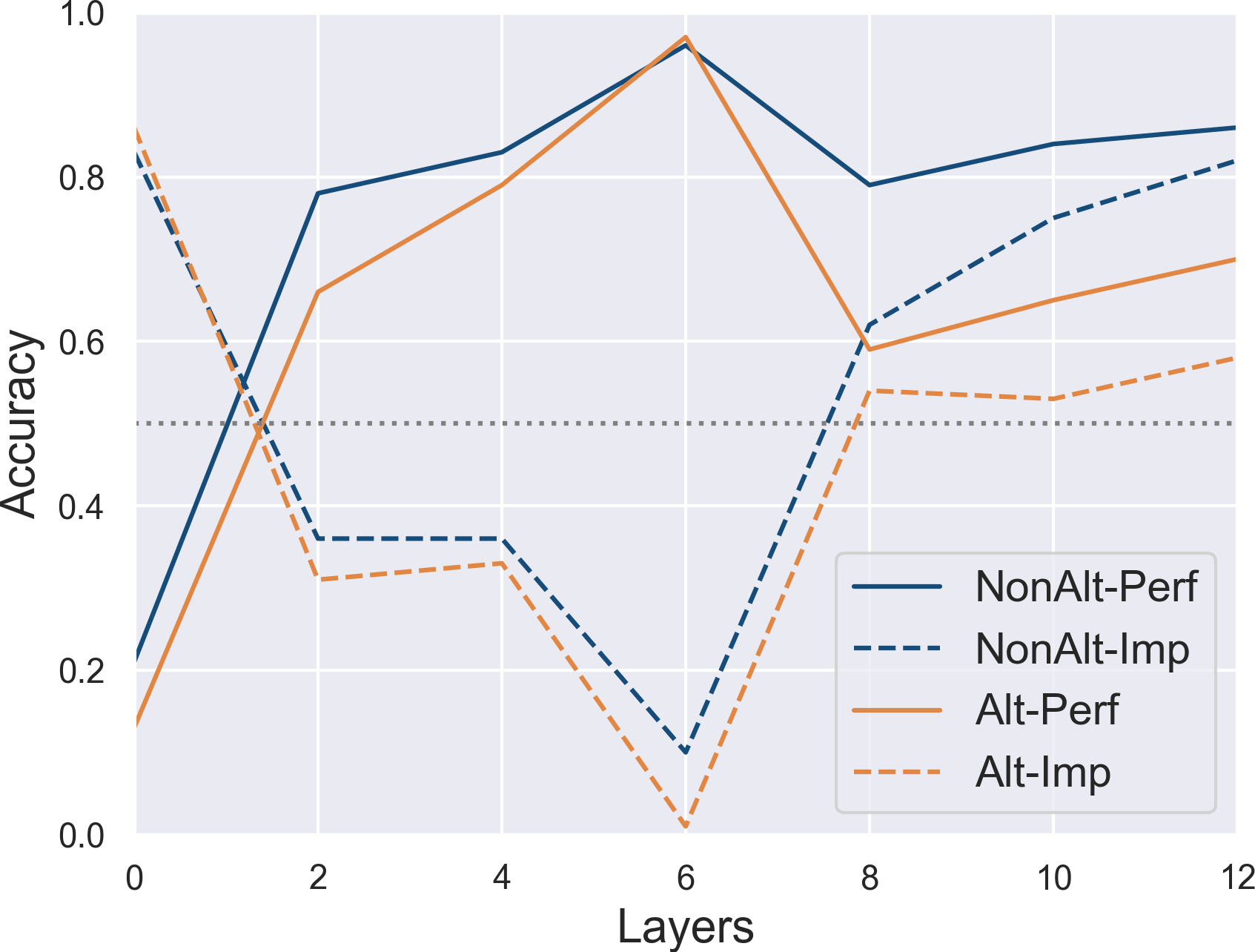}
\end{minipage}
\caption{Performance of BERT-base on iterative masking (left) and aspect inference (right) for target verbs. \textit{Perf} and \textit{Imp} denote perfective and imperfective aspect in non-alternative (\textit{NonAlt}) and alternative (\textit{Alt}) contexts.}
\label{fig:types-base}
\end{figure*}

For each target verb, we generated an aspect pair that differs only by the category of aspect. For this purpose, we used a list of over 2K verb lemmas with their aspect pairs which was manually created in collaboration with several linguists and Russian teaching experts. We generated an aspect pair for each target verb form automatically using a morphological generator which takes as input the verb lemma and a list of grammatical tags. For example, for a perfective verb form \textcyr{``получила''} (\textit{received}) in the past tense, singular number, and feminine gender, we generate an imperfective form \textcyr{``получала''}, which has the same tense, number, and gender. 

\begin{itemizerCompact}
    \item[] \textcyr{История \underline{получила} шикорое освещение в газетах.}
    \item[] \textit{The story \underline{received} extensive coverage in the newspapers.}
\end{itemizerCompact}

However, the paradigms of imperfective and perfective verb forms are not symmetrical. Perfective forms do not have present tense forms in the indicative mood, so we skip generation of an aspectual pair if the target verb is in imperfective present indicative form. There are no present tense forms for passive participles and transgressive forms, thus, in the context of this paper, we ignore participles and transgressives as targets.    

There is also a difference between future tense forms for perfective and imperfective: imperfective verbs have analytic forms, e.g., compare \textcyr{``прочитает''} (\textit{will read}, perf.) and \textcyr{``будет читать''} (\textit{will read}, imp.). We generate aspect pairs for future tense taking this difference into account.

\section{Behavioral Probing}
\label{sec:masking}

\paragraph{Iterative Masking} Algorithm~\ref{alg:iterative} demonstrates the process of iterative masking described in subsection~\ref{behavioral}. 

\begin{figure*}[t]
\center
  \includegraphics[scale=0.6]{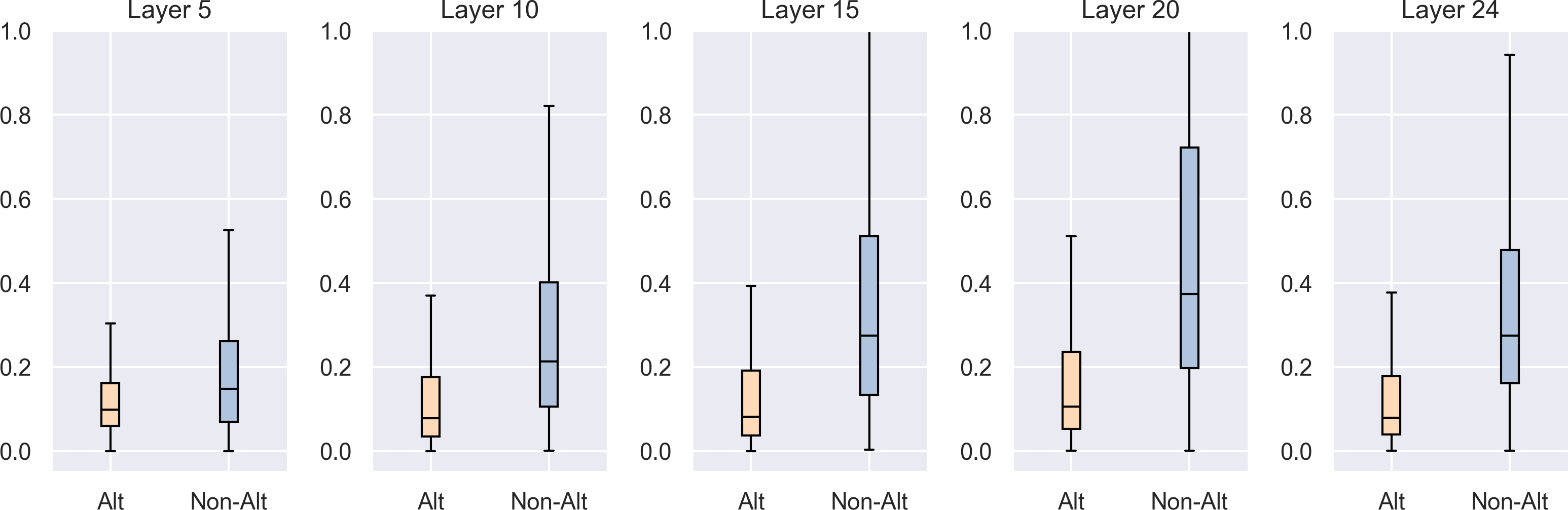}
  \caption{Difference between probabilities assigned by BERT-large to two aspect forms in alternative contexts (\textit{Alt}) vs. differences assigned to aspect forms in non-alternative contexts (\textit{Non-Alt}).}
  \label{fig:diff}
\end{figure*}

\begin{figure*}
\begin{minipage}{0.5\linewidth}
\centering
  \includegraphics[scale=0.5]{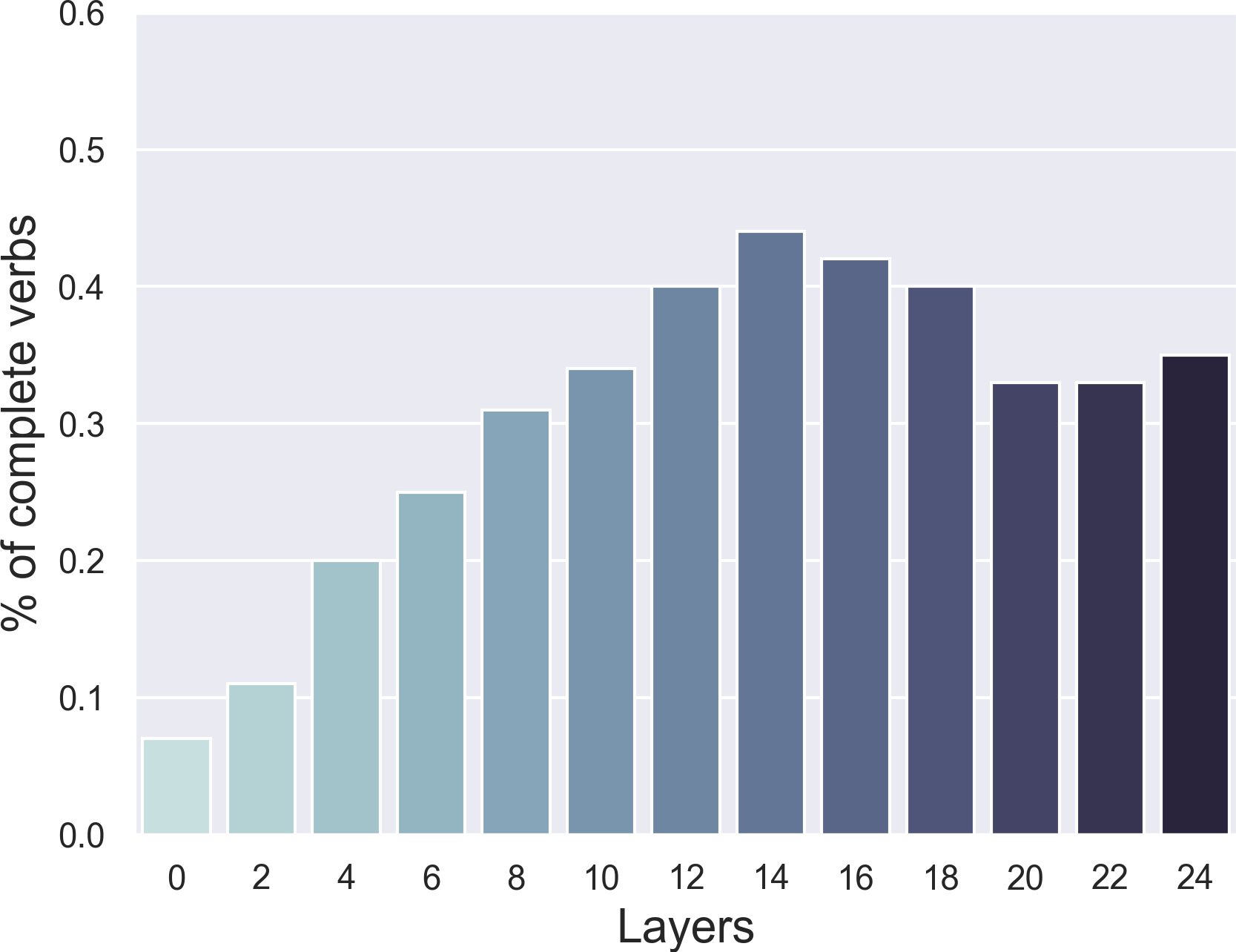}
\end{minipage}\hfill
\begin{minipage}{0.5\linewidth}
\centering
  \includegraphics[scale=0.5]{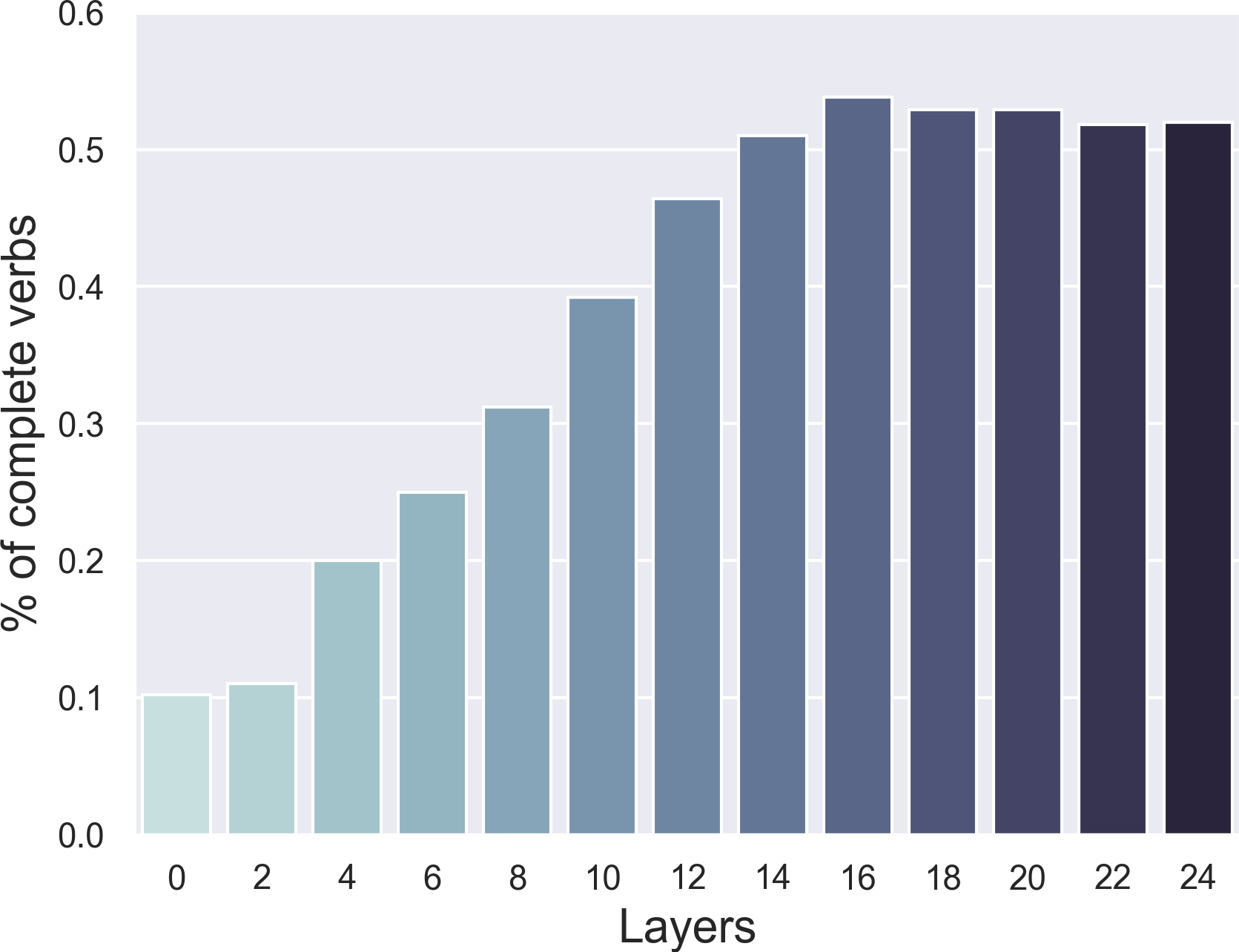}
\end{minipage}
\caption{Percent of tokens that are complete valid verb forms among top-k tokens for the masked position, k = 12000 (10\% of vocabulary size of BERT-large) on the left and k = 1200 (1\% of vocabulary size of BERT-large) on the right.}
\label{fig:complete-1}
\end{figure*}

Figure~\ref{fig:diff} shows boxplots with removed outliers displaying differences between probabilities assigned by BERT-large to two aspect forms (expected and complementary) in alternative contexts vs. differences between probabilities assigned to two aspect forms in non-alternative contexts. The probability of each form is calculated using iterative masking. Probability difference is calculated by subtracting the probability of the expected form from the probability of the complementary form: $P_{\{\text{exp.}\}} - P_{\{\text{compl.}\}}$.

\paragraph{Aspect Inference}

We inspected how many words out of the top-$k$ filled by BERT-large in the [MASK] position are complete verbs, see plots for $k=$1.2K and $k=$12K in Figure~\ref{fig:complete-1}. For the first 6 layers, the number of complete verb forms is low and most of them are imperfective, for any masked position. The model starts to predict perfective forms only after layer 4. 

Considering that early layers incorporate less context information~\cite{10.1162/tacl_a_00349}, a higher preference for imperfective can be caused by frequency differences between aspect forms in the BERT's training data. Since the data used for pre-training is not available to us, we compared form frequencies in the SynTagRus corpus~\cite{droganova2018data}.  Imperfective is indeed more frequent (55\% vs. 44\%) in SynTagRus.  However, these statistics characterize only one dataset.  The frequency of aspect forms can depend on the genre of texts in the corpus.  For example, legal texts usually have present tense more frequently than past or future.  As a result, imperfective forms dominate legal texts because present tense forms in the indicative mood do not exist for perfective in Russian. 

\begin{figure}[t]
\center
  \includegraphics[scale=0.5]{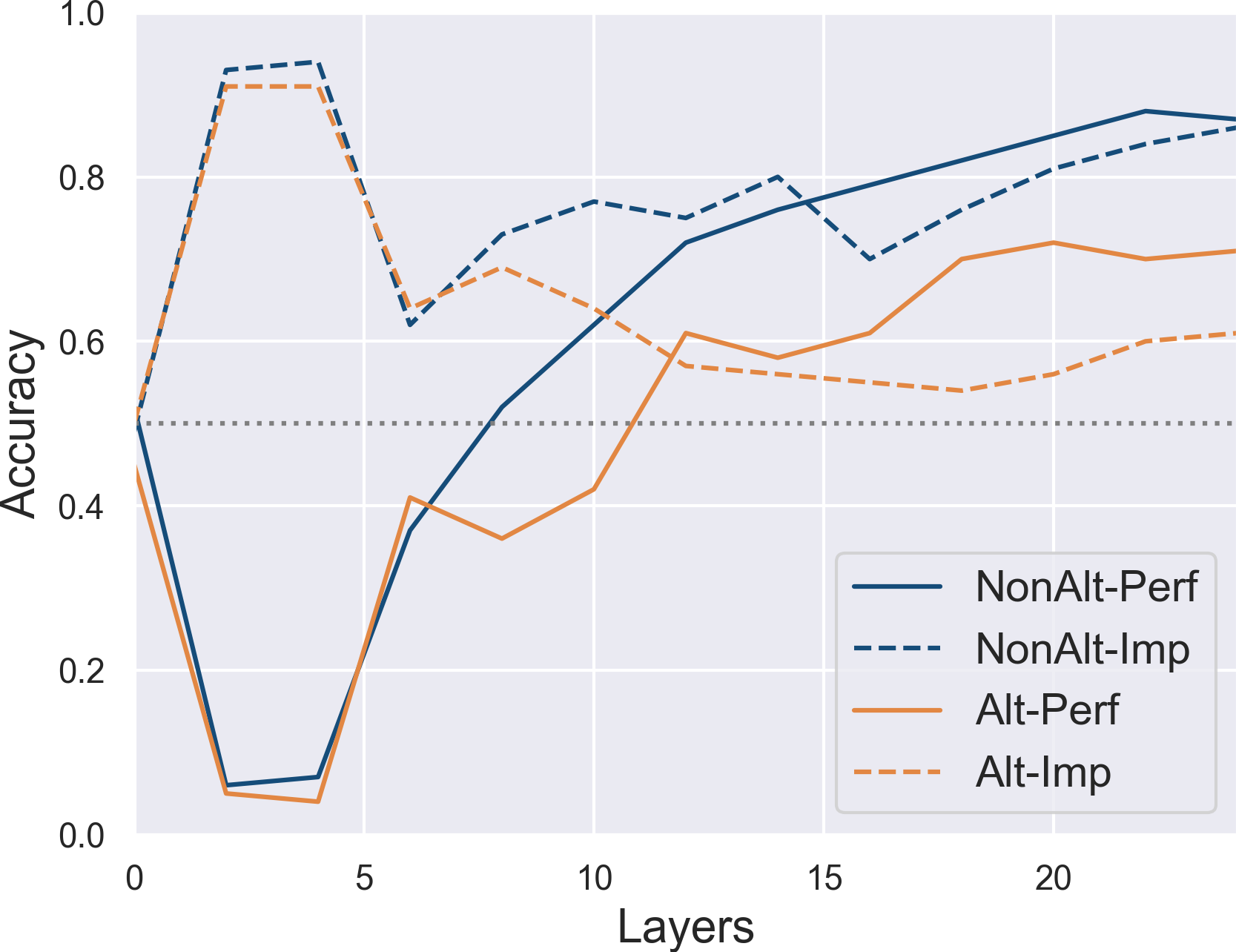}
  \caption{ Performance of BERT-large on aspect inference for the target verbs with k = 1200. Perf and Imp denote perfective and imperfective aspect in non-alternative (NonAlt) and alternative (Alt) contexts}
  \label{fig:large-1}
\end{figure}


\paragraph{BERT-base}

Figure~\ref{fig:types-base} shows the performance of BERT-base using iterative masking (left plot) and aspect inference (right plot) methods.




\begin{figure*}
\begin{minipage}{0.5\linewidth}
\centering
  \includegraphics[scale=0.55, trim=25mm 0mm 25mm 0mm ]{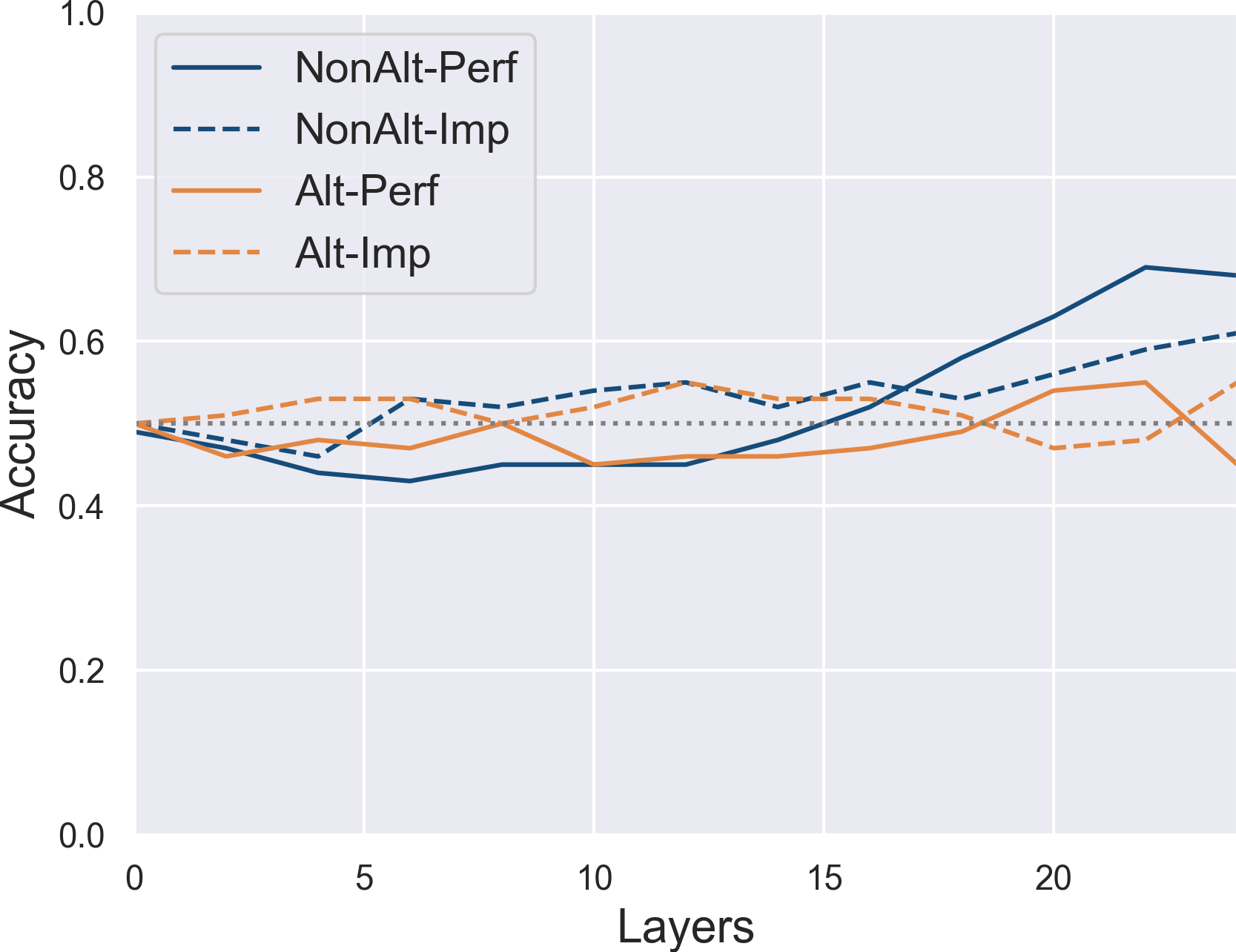}
\end{minipage}\hfill
\begin{minipage}{0.5\linewidth}
\centering
  \includegraphics[scale=0.55, trim=25mm 0mm 25mm 0 ]{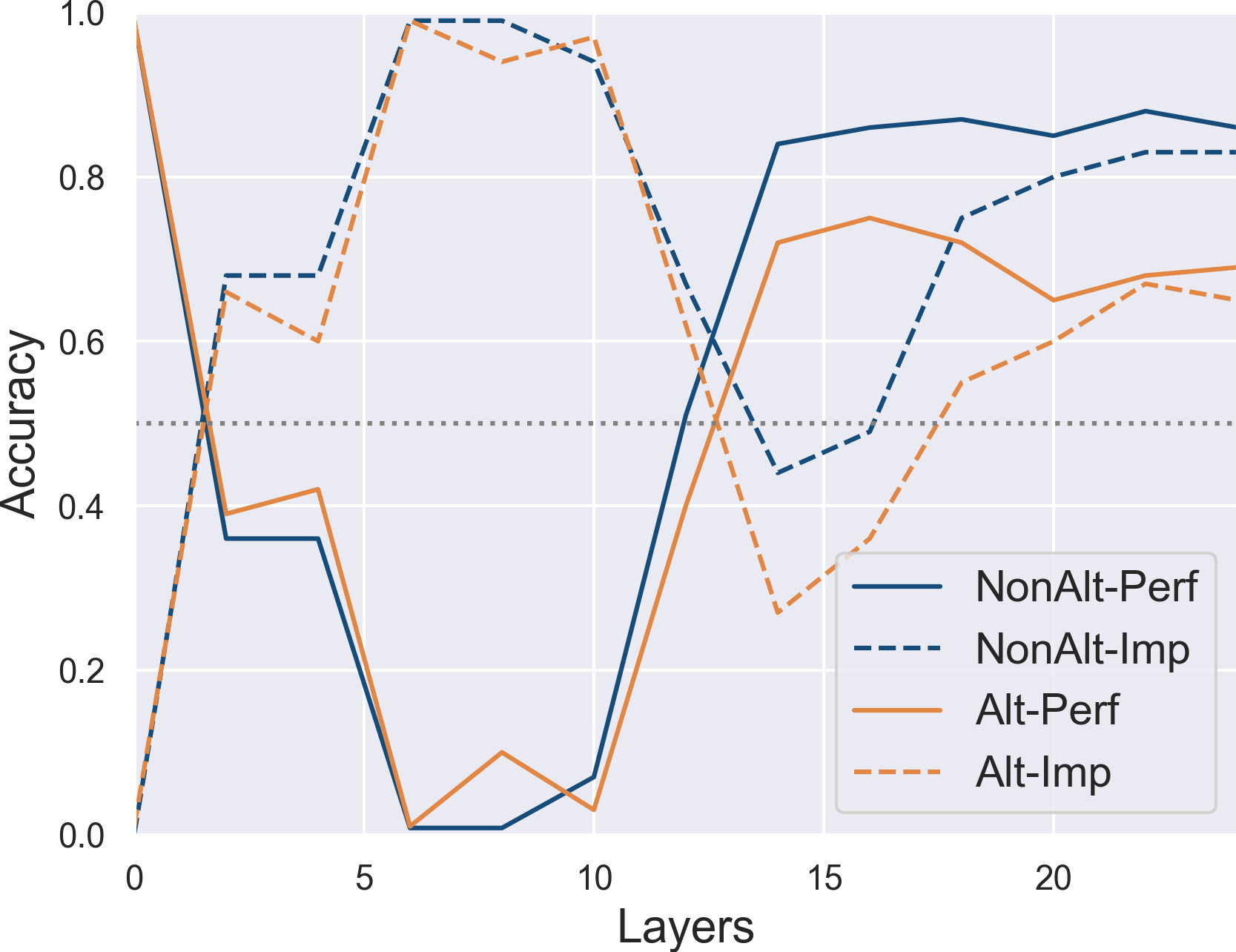}
\end{minipage}
\caption{Performance of RoBERTa-large on iterative masking (left) and aspect inference (right) for target verbs. \textit{Perf} and \textit{Imp} denote perfective and imperfective aspect in non-alternative (\textit{NonAlt}) and alternative (\textit{Alt}) contexts.}
\label{fig:types-roberta}
\end{figure*}


\section{Cue Words for Aspect}
\label{sec:dataset}

This section includes lists of lemmas of cue words that were used for collecting and annotating training data automatically for INLP classifiers. We excluded sentences where the target verb is negative since the negation particle ``\textcyr{не}'' (\textit{not}) in some contexts causes a change of aspect from perfective to imperfective, e.g., in the imperative mood.

Lemmas in brackets denote that any word in the first list appears with any word from the second list, e.g., \textcyr{``каждый день''} (\textit{every day}) or \textcyr{``каждый год''} (\textit{every year}). There is also a possibility for these words to be interrupted by their own dependent words, e.g., \textcyr{``каждый новый год''} (\textit{every new year}).

``Forbid'': \textcyr{[{запрещенный, дозволено, должен, надо, невозможно, нельзя, можно, нужно, обязан, опасный, рекомендуется, стоит}]}


``Iterative'': \textcyr{[бесконечно, бесперерывно, вечно, вновь, временами, всегда, часто, долго, изредка, непрерывно, как правило, постоянно, обычно, опять, регулярно, редко, систематически, снова]},                 
\noindent [\textcyr{[все, всякий, каждый, много, несколько, пара]} + \textcyr{[век, весна, вечер, вторник, воскресенье, год, день, десятилетие, зима, лето, месяц, миг, минута, неделя, ночь, осень, раз, сезон, секунда, среда, суббота, сутки, период, полдня, полночи, понедельник, пятница, четверг, утро, час]}],

\noindent [\textcyr{[по]} + \textcyr{[понедельник, вторник, среда, четверг, пятница, суббота, воскресенье, утро, вечер]}]

``Duration'': \textcyr{[[за]} + \textcyr{[век, весна, вечер, вторник, воскресенье, год, день, десятилетие, зима, лето, месяц, миг, минута, неделя, ночь, осень, раз, сезон, секунда, среда, суббота, сутки, период, полдня, полночи, понедельник, пятница, четверг, утро, час]]}

``Inception'': \textcyr{[браться, бросать, бросить, давать, взяться, заканчивать, закончить, кончить, надоедать, надоесть, начать, начинать, оканчивать, окончить, отвыкать, отвыкнуть, передумать, передумывать, переставать, перестать, приниматься, приняться, продолжать, продолжить, раздумать, раздумывать, разучиваться, разучиться, расхотеться, становиться, стать, уставать, устать]} 

``Like'': \textcyr{[запрещать, запрещаться, избегать, любить, научить, научиться, нравиться, отговатьвать, привыкнуть, привыкать, следовать, уметь, учиться]}

\noindent ``Forget'': \textcyr{[договариваться, договориться,  забывать, забыть, обещать, согласиться, соглашаться, удасться, успевать, успеть]}

``Capability'': \textcyr{[мочь, смочь, способный]}

``Result'': \textcyr{[вдруг, внезапно, наконец, уже]}

[\textcyr{[в]} + \textcyr{[итог, конец, результат, финал]} ]

\begin{figure*}[t!]
\begin{minipage}{0.5\linewidth}
\centering
  \includegraphics[scale=0.48, trim=25mm 0mm 25mm 0mm ]{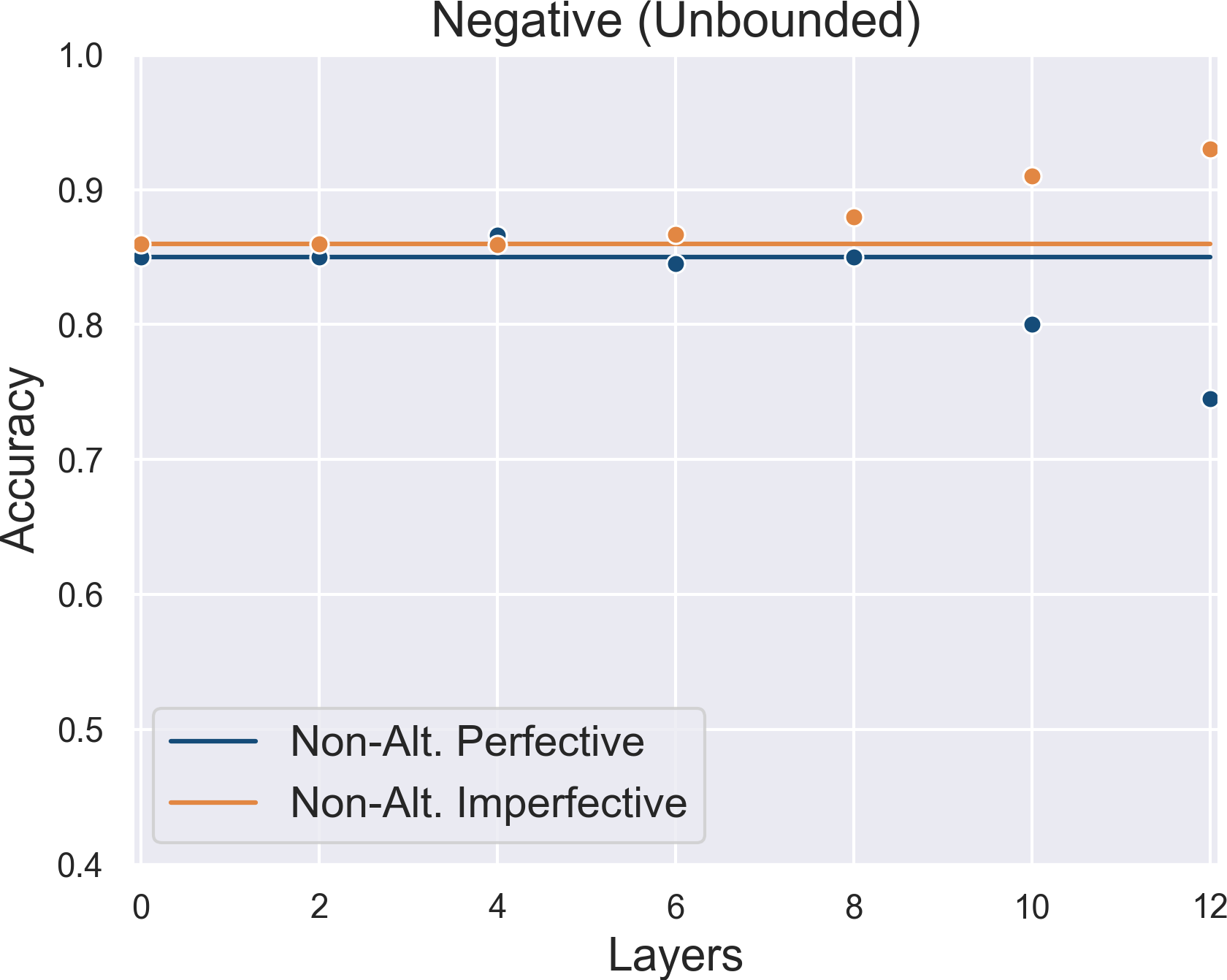}
\end{minipage}\hfill
\begin{minipage}{0.5\linewidth}
\centering
  \includegraphics[scale=0.48, trim=0mm 0mm 0mm 0 ]{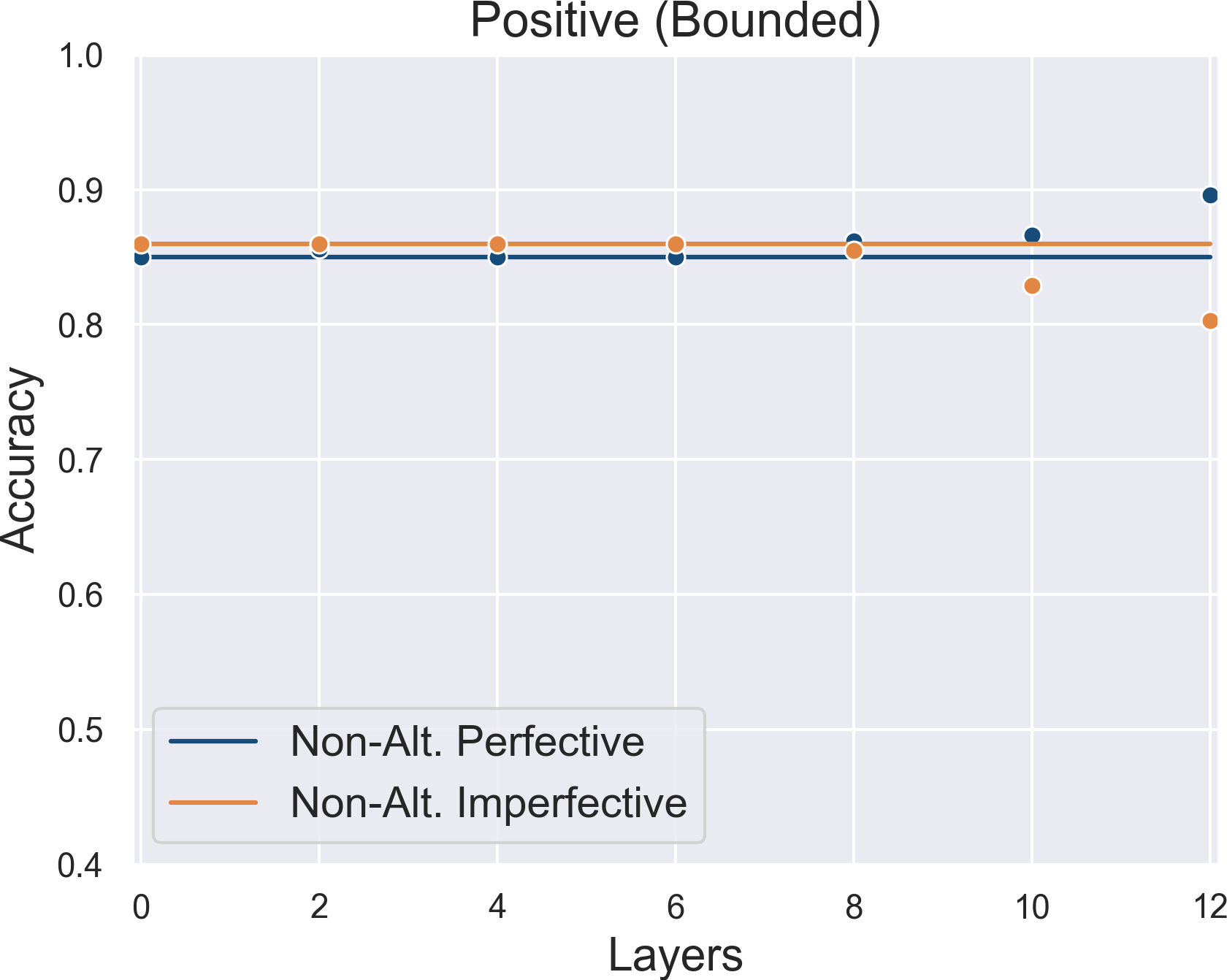}
\end{minipage}
\begin{minipage}{0.5\linewidth}
\centering
  \includegraphics[scale=0.48, trim=0mm 0mm 0mm 0 ]{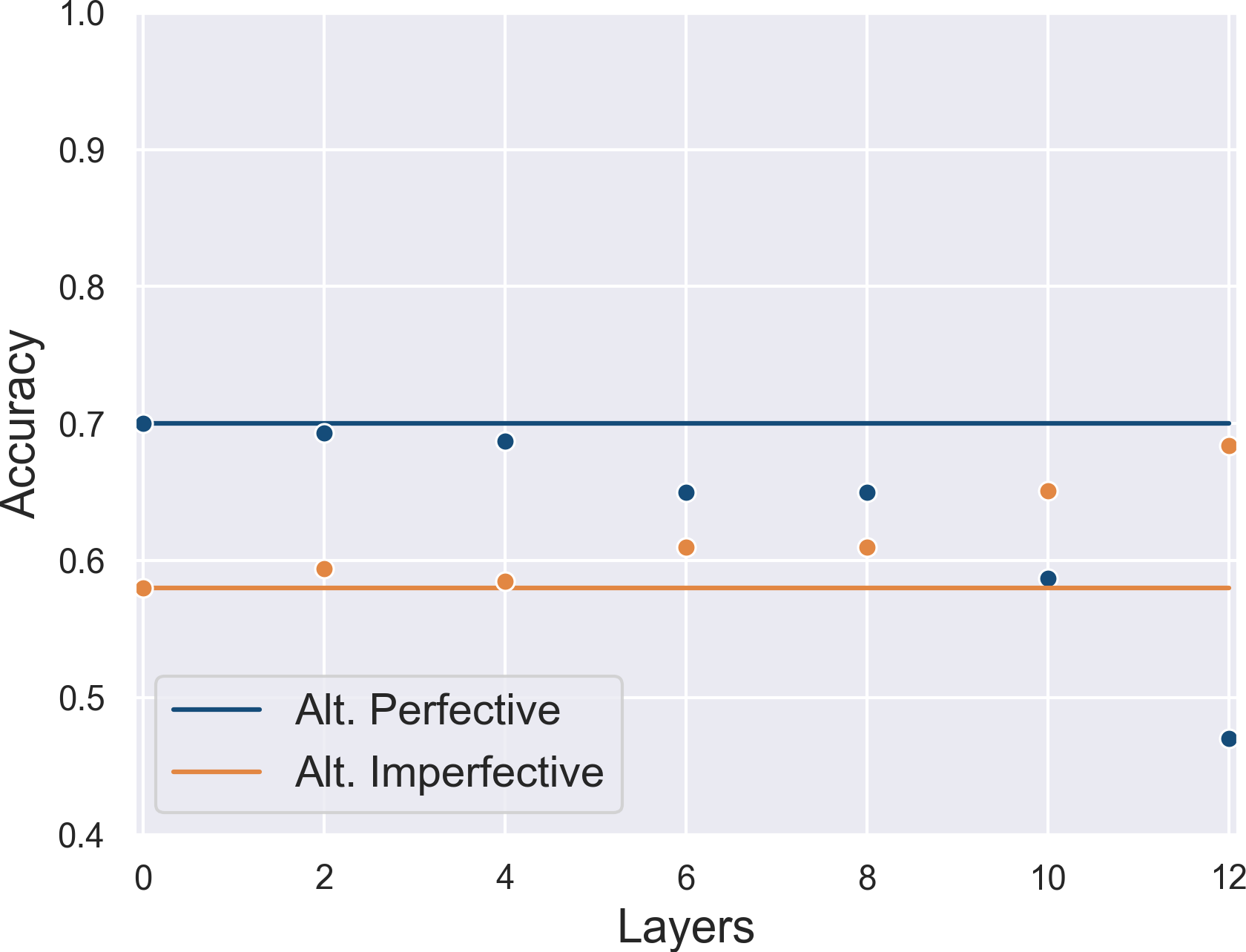}
\end{minipage}\hfill
\begin{minipage}{0.5\linewidth}
\centering
  \includegraphics[scale=0.48, trim=0mm 0mm 0mm 0 ]{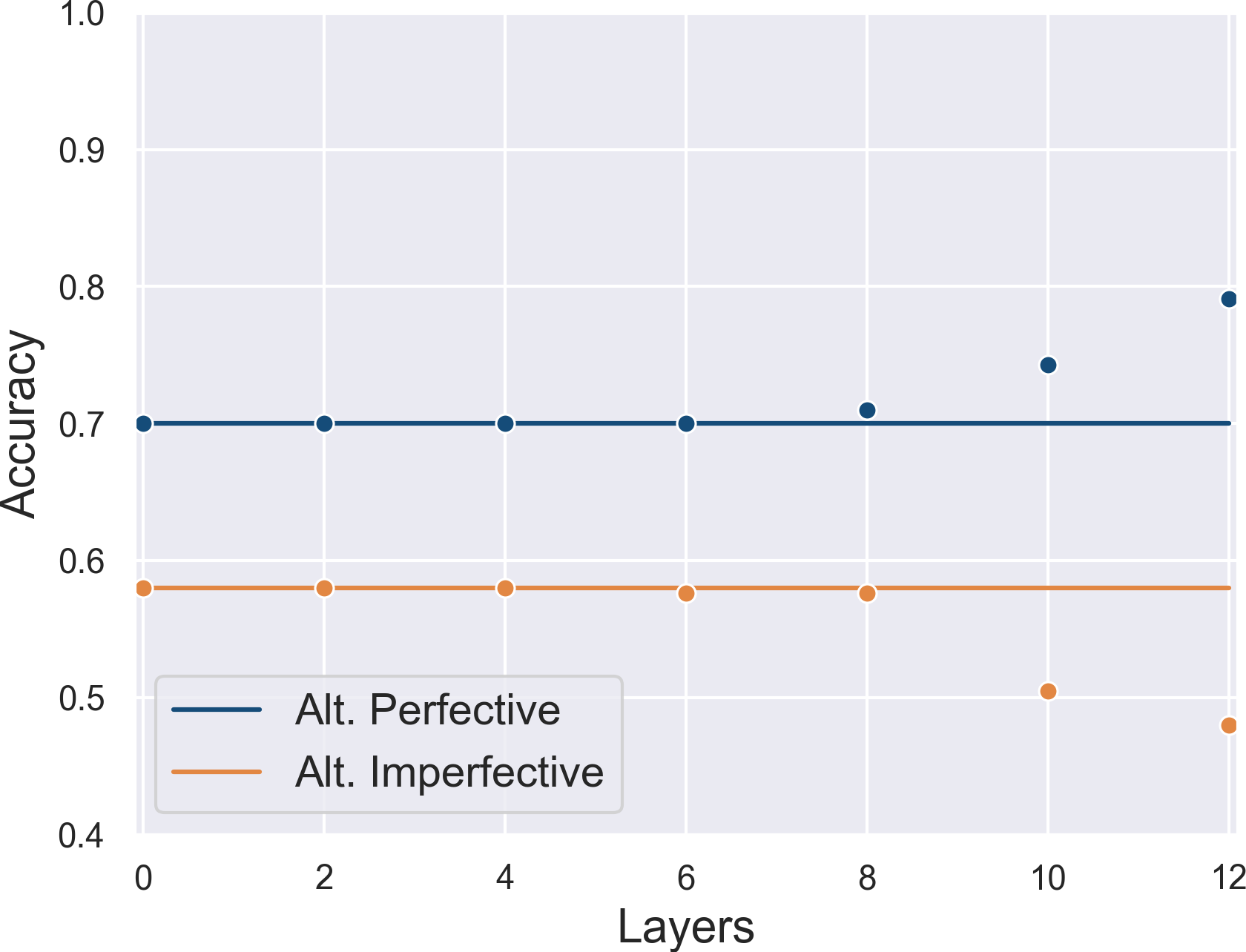}
\end{minipage}
\caption{Change in accuracy of predicting correct (expected) aspect using aspect inference method after interventions on BERT-base representations. Top plots show results in non-alternative contexts; bottom plots---in alternative contexts. Left plots show the results of negative interventions: moving toward the meaning of unbounded action. Right plots---results of positive interventions: moving toward the meaning of bounded action. \textbf{Flat} lines indicate performance before interventions. \textbf{Dots}---after interventions. \textbf{Dashed}---after random interventions.} 
\label{types-base}
\end{figure*}

\begin{figure*}[t!]
\begin{minipage}{0.5\linewidth}
\centering
  \includegraphics[scale=0.48, trim=0mm 0mm 0mm 0 ]{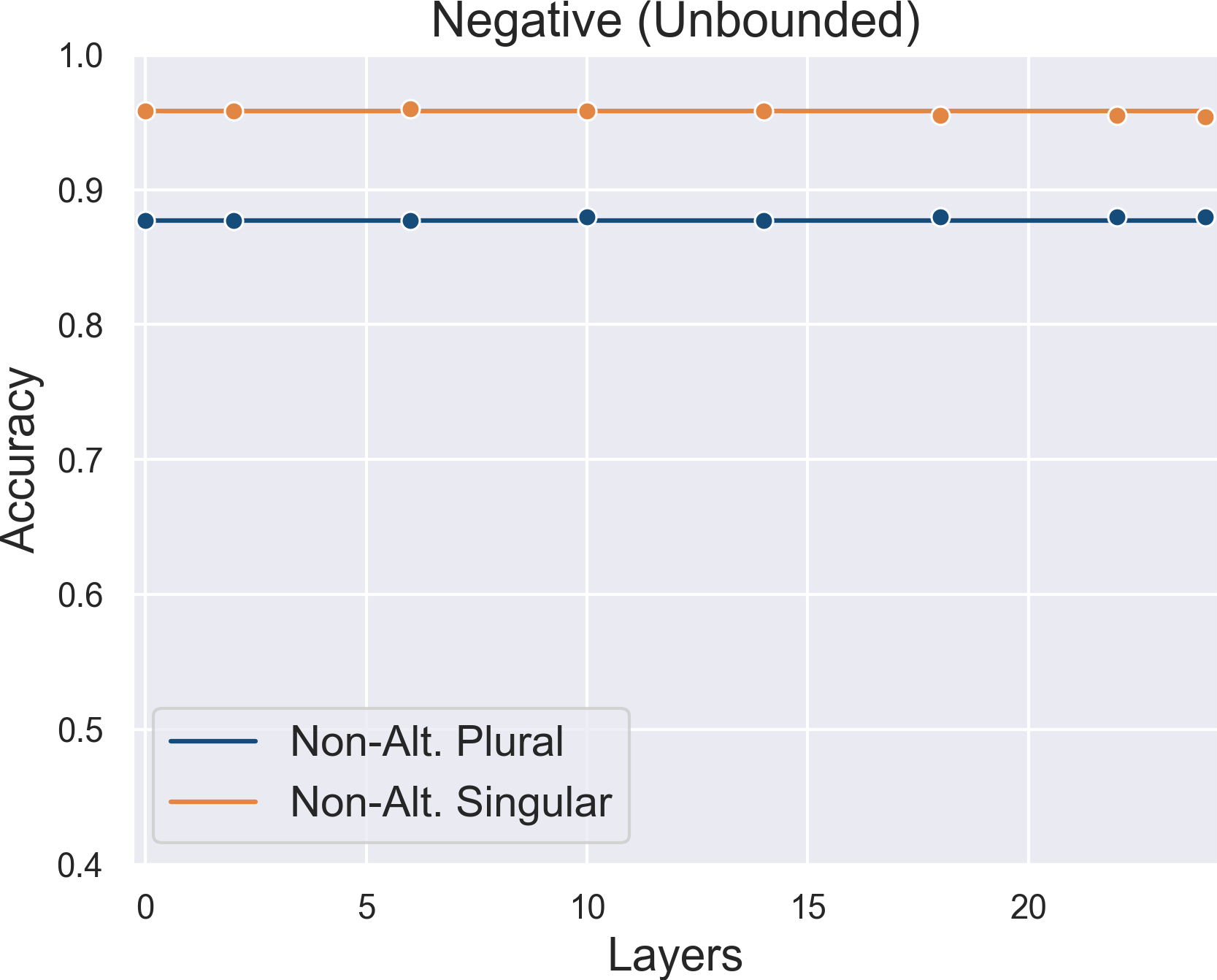}
\end{minipage}
\begin{minipage}{0.5\linewidth}
\centering
  \includegraphics[scale=0.48, trim=0mm 0mm 0mm 0 ]{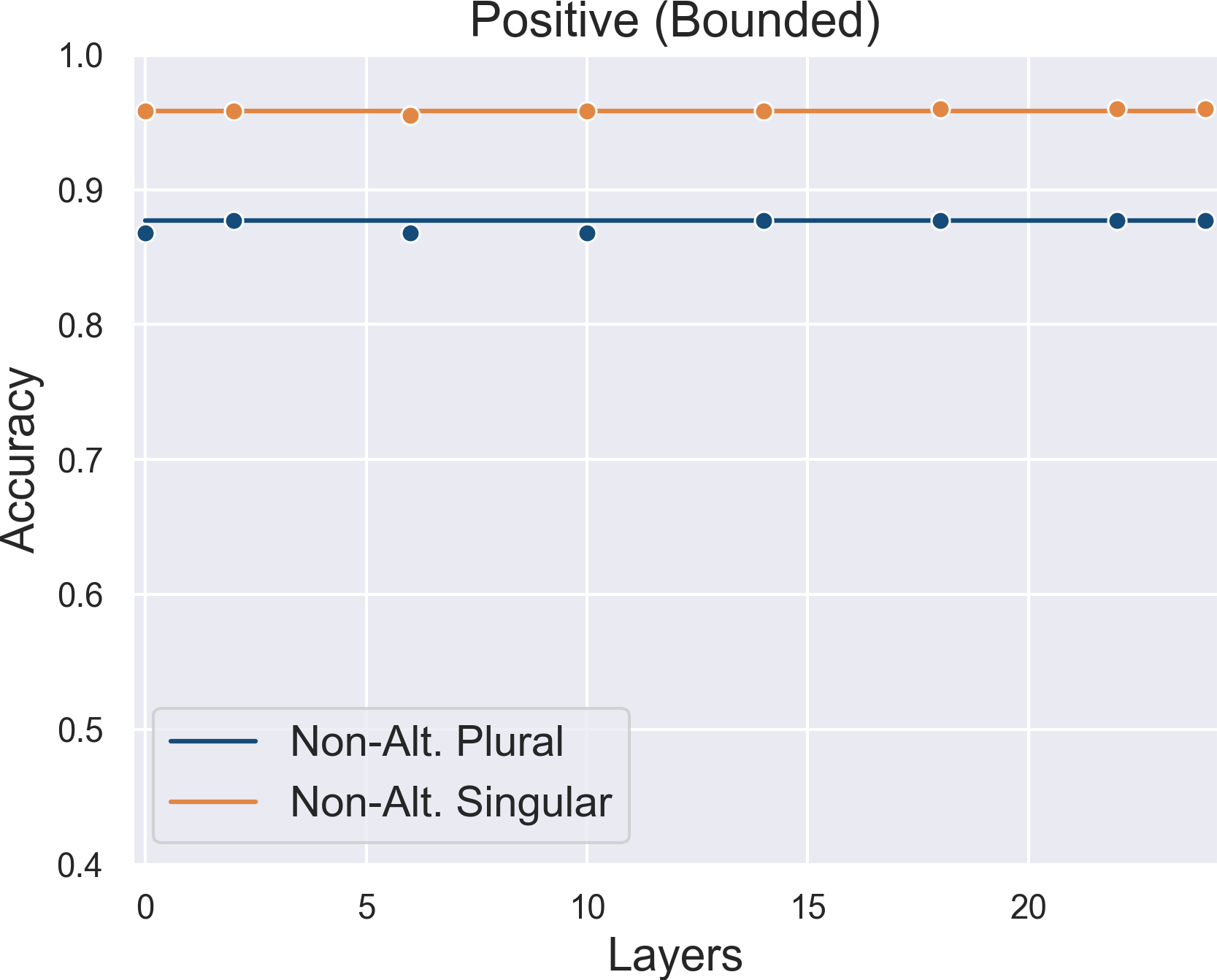}
\end{minipage}
\begin{minipage}{0.5\linewidth}
\centering
  \includegraphics[scale=0.48, trim=0mm 0mm 0mm 0 ]{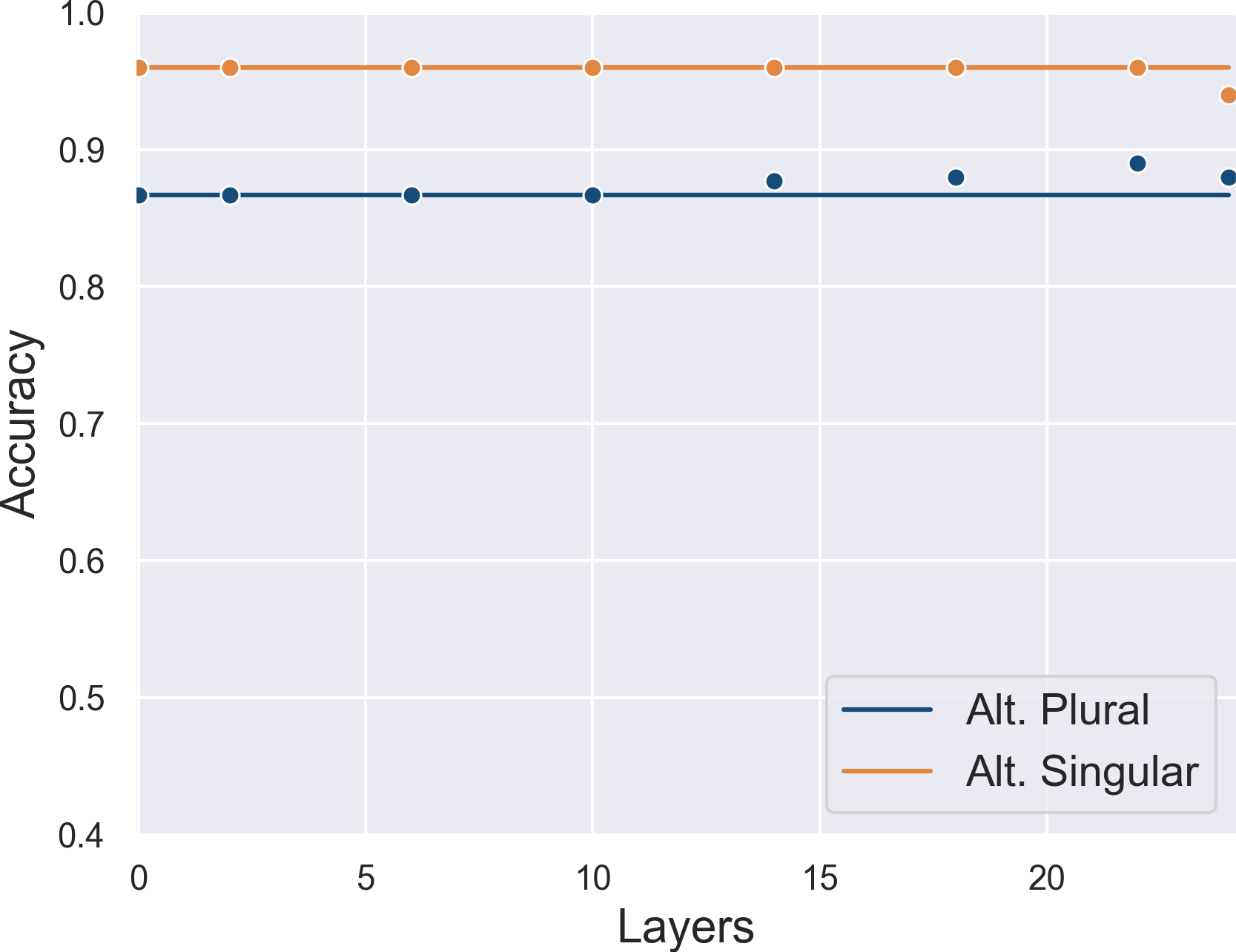}
\end{minipage}
\begin{minipage}{0.5\linewidth}
\centering
  \includegraphics[scale=0.48, trim=0mm 0mm 0mm 0 ]{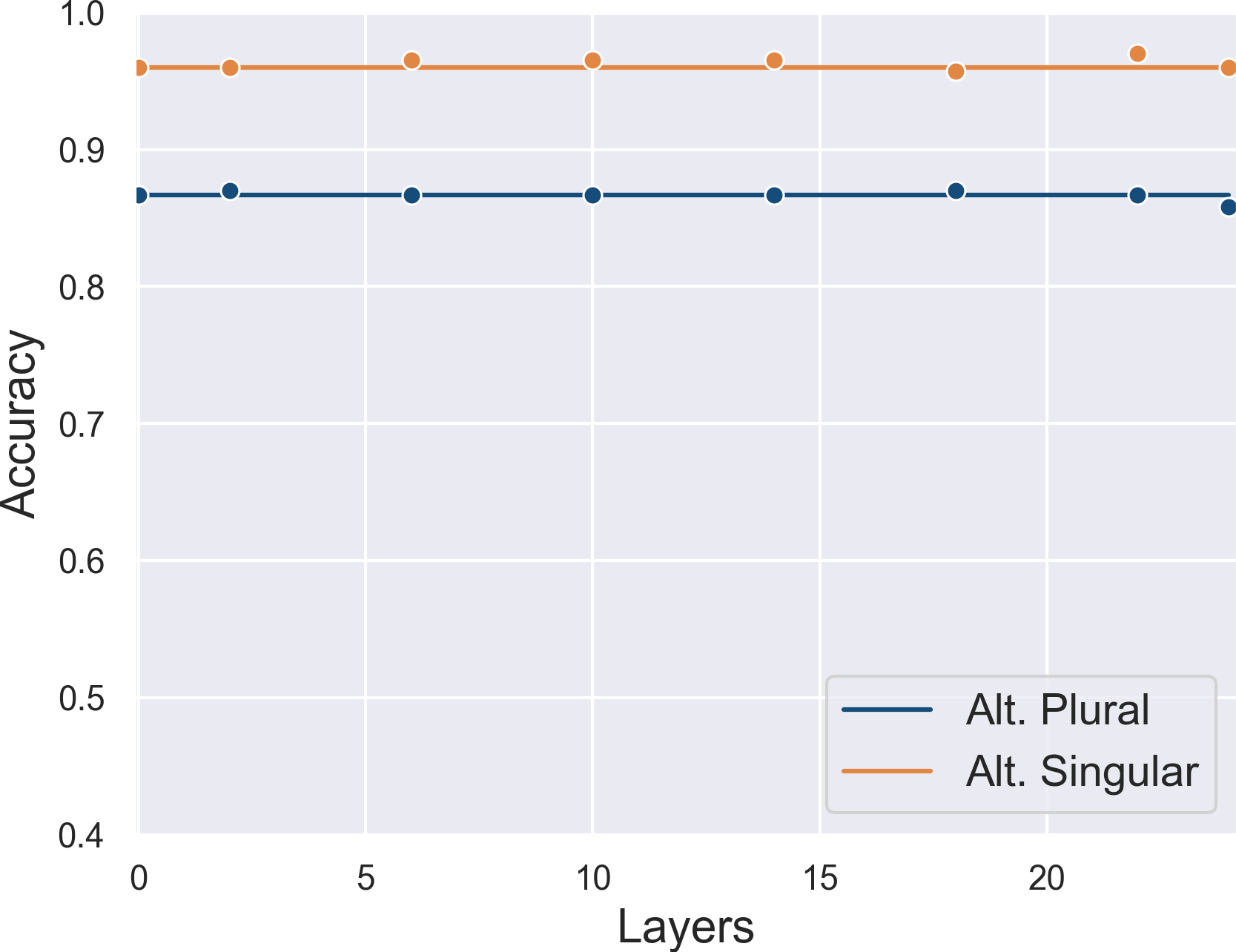}
\end{minipage}
\caption{Change in accuracy of predicting correct number using the number inference after interventions on BERT-large representations. Top plots show results in non-alternative contexts; bottom plots---in alternative contexts. Left plots show the results of negative interventions: moving toward the meaning of unbounded action. Right plots---results of positive interventions: moving toward the meaning of bounded action. \textbf{Flat} lines indicate performance before interventions. \textbf{Dots}---after interventions.}
\label{fig:number}
\end{figure*}

\begin{figure*}[t!]
\begin{minipage}{0.5\linewidth}
\centering
  \includegraphics[scale=0.48, trim=0mm 0mm 0mm 0 ]{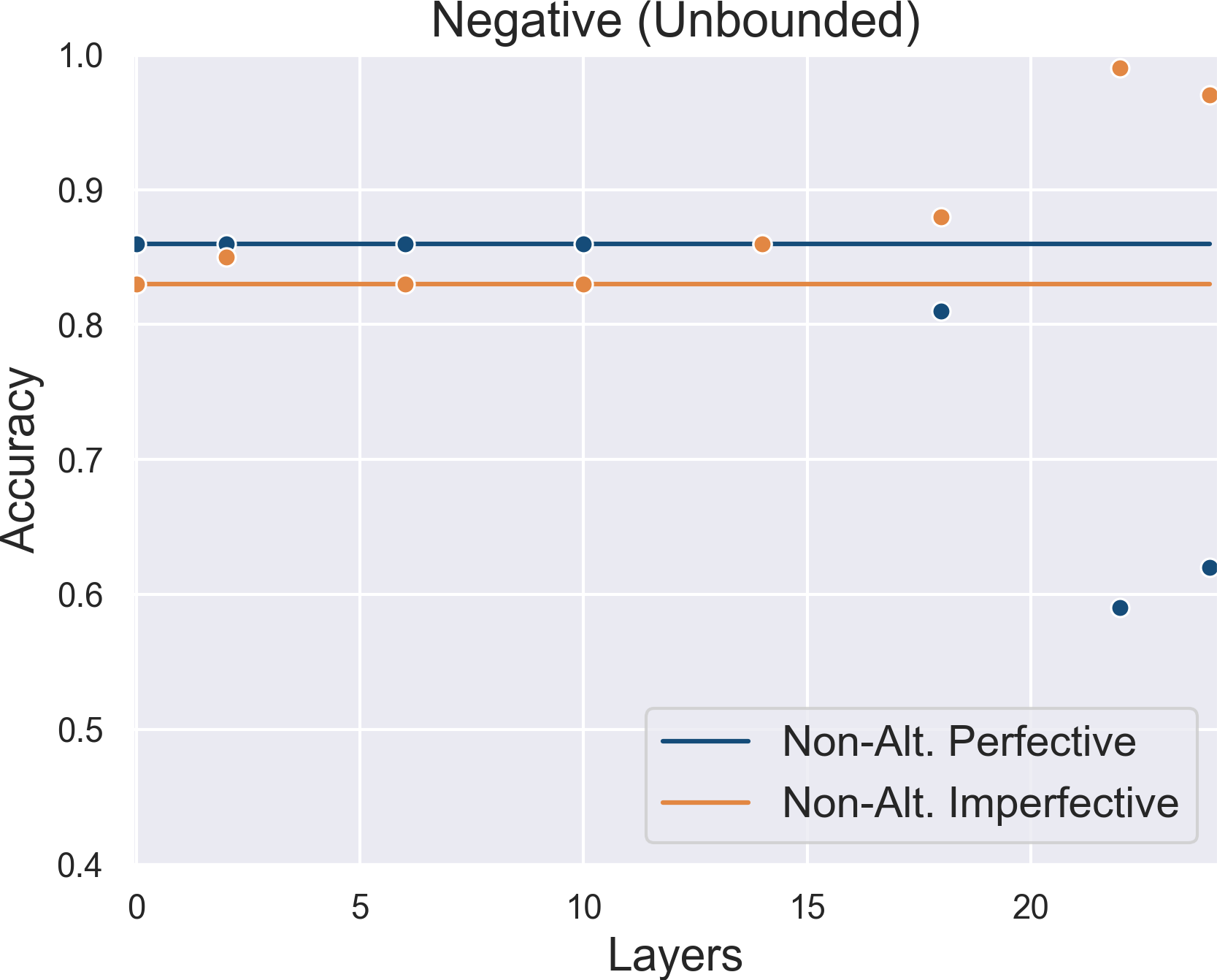}
\end{minipage}
\begin{minipage}{0.5\linewidth}
\centering
  \includegraphics[scale=0.48, trim=0mm 0mm 0mm 0 ]{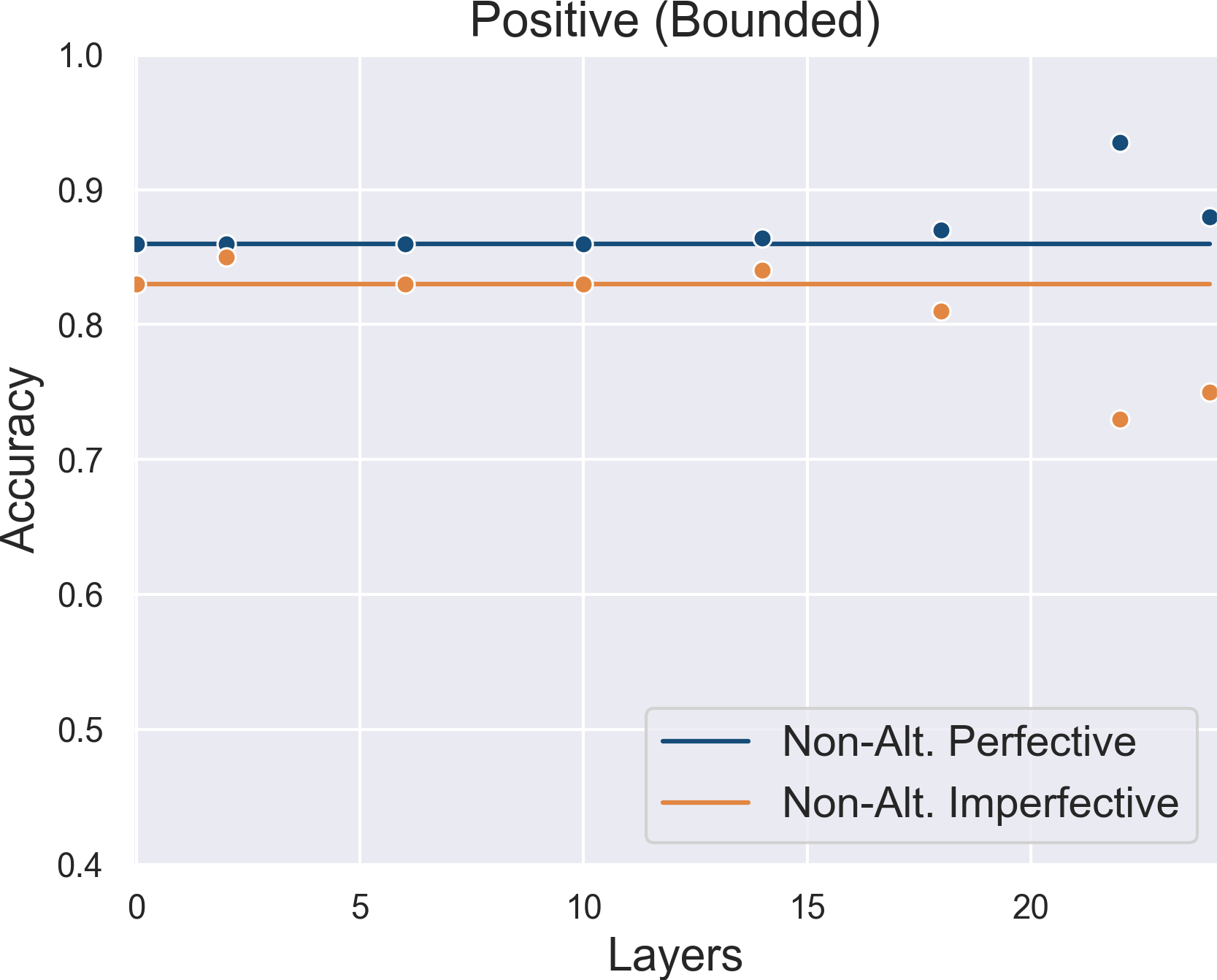}
\end{minipage}
\begin{minipage}{0.5\linewidth}
\centering
  \includegraphics[scale=0.48, trim=0mm 0mm 0mm 0 ]{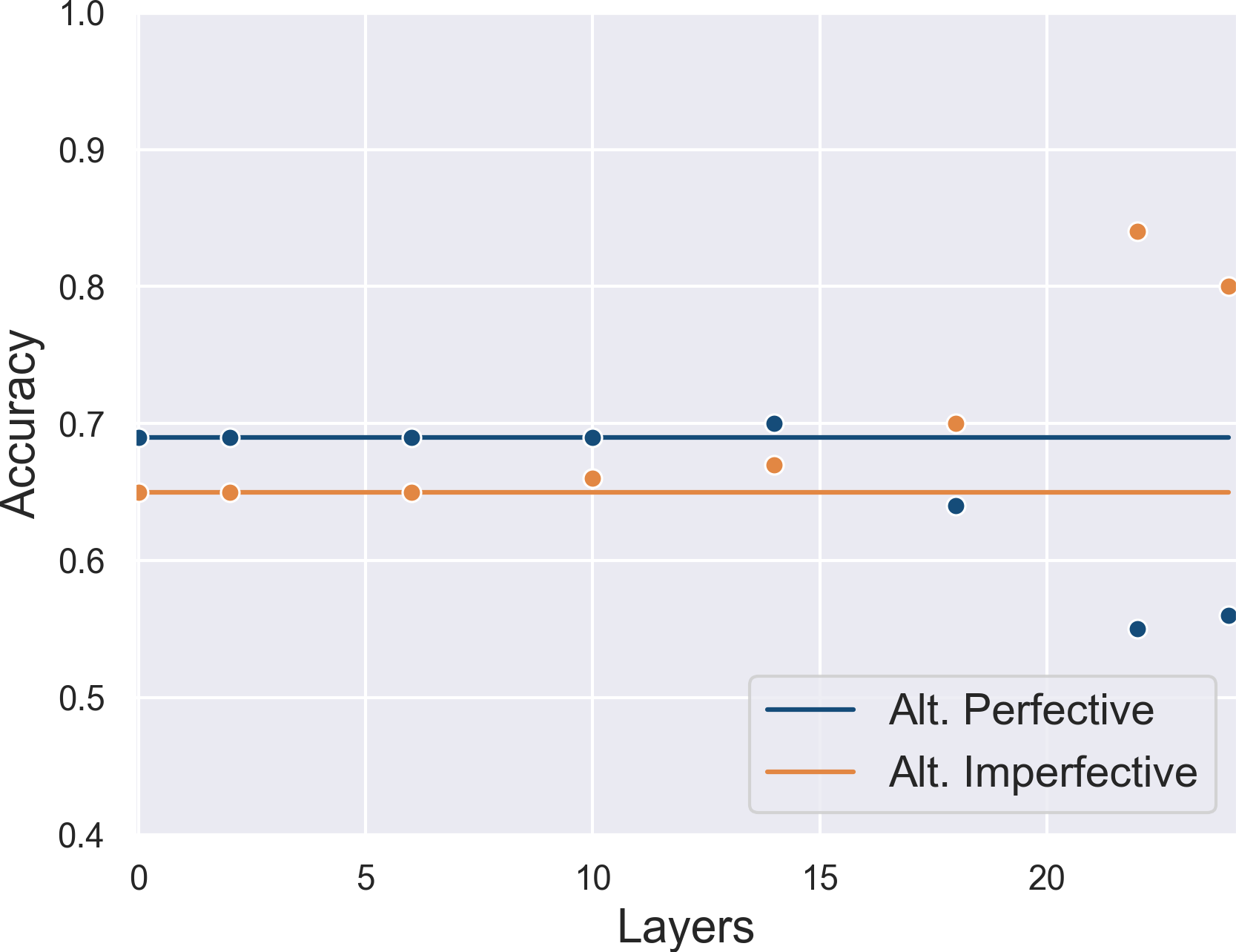}
\end{minipage}
\begin{minipage}{0.5\linewidth}
\centering
  \includegraphics[scale=0.48, trim=0mm 0mm 0mm 0 ]{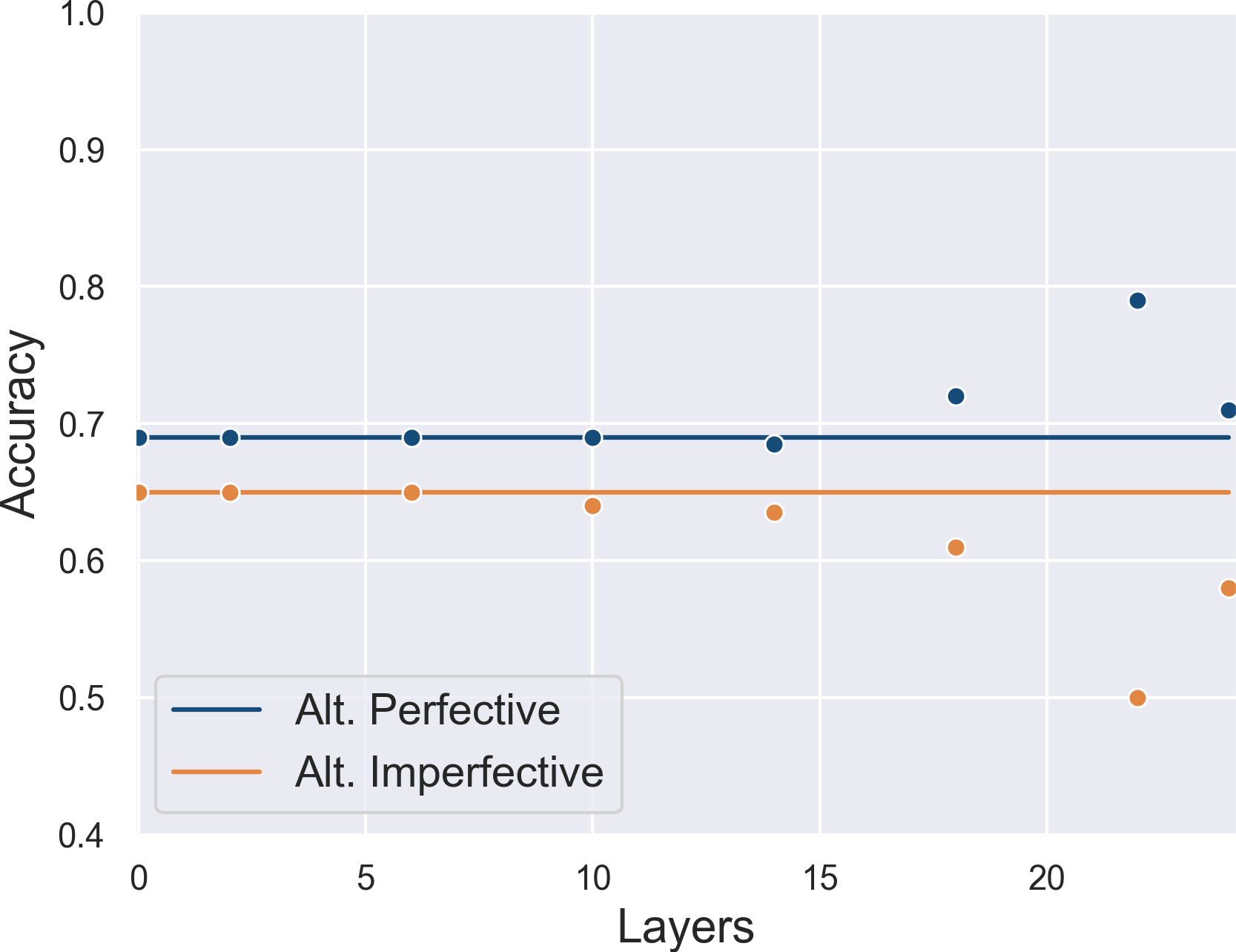}
\end{minipage}
\caption{Change in accuracy of predicting correct (expected) aspect using the aspect inference after interventions on RoBERTa-large representations. Top plots show results in non-alternative contexts; bottom plots---in alternative contexts. Left plots show the results of negative interventions: moving toward the meaning of unbounded action. Right plots---results of positive interventions: moving toward the meaning of bounded action. \textbf{Flat} lines indicate performance before interventions. \textbf{Dots}---after interventions.}
\label{fig:inlp-roberta}
\end{figure*}

\section{Training INLP}
\label{app:inlp}

We use SVM with stochastic gradient descent learning\footnote{\href{https://scikit-learn.org/stable/modules/generated/sklearn.linear\_model.SGDClassifier.html}{sklearn.linear\_model.SGDClassifier}} as an INLP classifier and set the number of classifiers $m = 20$ and $\alpha = 4$ for BERT-large. \citet{ravfogel-etal-2021-counterfactual} demonstrate that using different parameter values $m$ and $\alpha$ does not substantially affect observed results. We use $m = 10$ for BERT-base. We increased $m$ for BERT-large since its hidden representations are twice as big. Other parameters of the INLP classifier are: adaptive learning rate, early stopping set to True, and eta=0.1.

\section{Effect of Counterfactuals on Aspect}
\label{app:base}

Figure~\ref{fig:types-base} shows the effect of counterfactual interventions on predicting aspect of the target verb using BERT-base. The effects of positive and negative interventions in both alternative and non-alternative contexts are similar to those observed using BERT-large. Interventions into the boundedness of action have a bigger impact on predicting aspect in alternative contexts. Positive and negative interventions affect aspect prediction in the last layers of BERT-base as well, predominantly after layer 8.

\section{Affect of Boundedness on Category of Number}
\label{sec:number}

Figure~\ref{fig:number} shows the effect of counterfactual interventions on predicting the number of the target verb. Interventions affect the meaning of the boundedness of the described action: whether the action is bounded or unbounded. Plots demonstrate that predicting the category of number is not affected by altering boundedness of the action, unlike aspect.

\begin{table}[t!]
  \begin{center}
      \scalebox{0.80}{
  \begin{tabular}{l|cc|cc}
\hline
    \multicolumn{1}{l|}{} & \multicolumn{2}{c|}{non-alternative}  & \multicolumn{2}{c}{alternative} \\
      Model & $F_{0.5}^{\text{perf}}$  & $F_{0.5}^{\text{imp}}$ & $F_{0.5}^{\text{perf}}$  & $F_{0.5}^{\text{imp}}$ \\\hline
    Pretrained & 36.3 & 49.2 & 54.0 & 51.1	 \\
    Fine-tuned & 85.9 & 84.0 & 67.5 & 57.0  \\
    Fine-tuned (up to last 5)  & 85.0 & 83.9 & 67.0 & 56.0 \\  
    Fine-tuned (last 6) & 87.0 & 88.0 & 69.0 & 64.0 	\\
    Fine-tuned (last 5) &\textbf{88.5} & \textbf{88.0} & \textbf{69.1}  & \textbf{64.2} \\
    Fine-tuned (last 2) & 87.0 & 87.0 & 67.0 & 60.0 	\\
    Fine-tuned (last 1) & 86.0 & 87.0 &	67.0 & 60.0 \\
\hline
  \end{tabular}
  }
  \caption{Performance in terms of $F_{0.5}$ for imperfective and perfective aspects in non-alternative (non-alt.) and alternative (alt.) contexts.}
  \label{table:perf-full}
  \end{center}
\end{table}

\section{Fine-tuning BERT for Aspect Prediction}
\label{sec:finetune}

Training data for fine-tuning BERT was generated using the SynTagRus corpus. For every sentence, we picked all verbs, labeled them with their aspect (\texttt{Perf} or \texttt{Imp} tag in the morphological analysis), and replaced them with a [MASK] token; all other words were labeled with \texttt{None}. Masking was used because the task is not to predict an aspect of a given verb form, but to predict which aspect fits in the given context. Also, during inference, we do not know which form should fit the context. We generated 60K training sentences and 7.5K validation sentences, where each sentence includes two masked verbs on average.

Parameters of training: learning rate = 5e-5, epochs = 3, batch size = 256, max input length = 512. The model was fine-tuned using 2 GPUs NVIDIA A100.

Testing was performed using the same data that we used for all probing tasks. The fine-tuned model is successful if the predicted label is the same as the expected aspect of the target verb. Table~\ref{table:perf} and Table~\ref{table:perf-full} report results averaged across 5 runs for each model configuration.

\end{document}